\documentclass{article} 
\usepackage{iclr2027_conference,times}
\usepackage[T1]{fontenc}

\usepackage{amsmath,amsfonts,bm}

\def\eqref#1{equation~\ref{#1}}

\def\1{\bm{1}}

\DeclareMathAlphabet{\mathsfit}{\encodingdefault}{\sfdefault}{m}{sl}
\SetMathAlphabet{\mathsfit}{bold}{\encodingdefault}{\sfdefault}{bx}{n}

\usepackage{hyperref}
\usepackage{url}
\usepackage{booktabs}
\usepackage{multirow}
\usepackage{amsmath}
\usepackage{graphicx}
\usepackage{wrapfig}
\usepackage{adjustbox}
\usepackage{arydshln}
\usepackage{subcaption}
\usepackage{caption}
\usepackage{tabularx}
\usepackage{array}
\usepackage{makecell}
\usepackage{xcolor}

\title{MASKerade: Token-Routed Mask Experts for Dense-to-MoE Upcycling}

\author{
\begin{tabular}{@{}l@{}}
\textbf{Mingyuan Zhang$^{1}$, Yue Bai$^{1}$, Zhongruo Wang, Yupin Huang,}\\
\textbf{Yiyang Huang$^{1}$, Hailing Wang$^{1}$, Huimin Zeng$^{1}$, Yun Fu$^{1,2}$}
\end{tabular}\\[4pt]
$^{1}$Department of Electrical and Computer Engineering, Northeastern University\\
$^{2}$Khoury College of Computer Science, Northeastern University
}

\newcommand{\ms}[2]{#1{\scriptsize$\pm$#2}} 

\iclrfinalcopy 
\begin{document}

\maketitle
\lhead{}
\renewcommand{\headrulewidth}{0pt}

\begin{abstract}
Sparsely activated Mixture-of-Experts (MoE) models increase model capacity without a proportional increase in per-token computation. Dense-to-MoE upcycling reuses pretrained dense models to construct such systems, commonly by copying feed-forward networks (FFNs) into independently trained experts. We introduce \textbf{MASKerade}, a dense-to-MoE training method that instead learns experts as sparse subnetworks of a frozen pretrained FFN. Each expert is defined by a learned binary mask, and a token-level router selects which masked FFNs to execute and combine. The router and mask scores are optimized jointly, while the underlying FFN weight values remain unchanged. This formulation supports neuron-structured, semi-structured, and unstructured experts within the same routing architecture. Our main configuration uses four 2:4 experts with top-2 routing, where two half-dense expert passes have the nominal FFN arithmetic of one dense pass, without requiring independent expert weight matrices. On five vision-language benchmarks with Qwen and Gemma backbones, this configuration achieves the highest performance among the compared baselines. Comparisons across mask granularities, routing interventions, and compute-matched controls distinguish the effects of learned connectivity from expert activation count. These results establish mask learning over frozen weights as a practical alternative for constructing token-routed MoE experts. Codes: \emph{\href{https://github.com/Ming-K9/MASKerade}
{\textcolor{magenta}{https://github.com/Ming-K9/MASKerade}}}.
\end{abstract}

\section{Introduction}
\label{sec:introduction}

Mixture-of-Experts (MoE) models separate the number of available parameters from per-token computation~\citep {jacobs1991adaptive}.
A sparsely gated layer selects a small subset of experts and combines their outputs, rather than evaluating every expert for every input~\citep{shazeer2017outrageouslylargeneuralnetworks,fedus2022switchtransformersscalingtrillion}.
Dense-to-MoE upcycling makes this approach accessible from existing pretrained models by converting selected feed-forward networks (FFNs) into routed expert layers, reusing the investment in dense pretraining~\citep{komatsuzaki2023sparse}.
This is particularly attractive for multimodal instruction tuning, where a single model must process heterogeneous visual and textual inputs~\citep{11353922,chen2024llavamolesparsemixturelora}.

An important design decision in upcycling is \emph{what constitutes an expert}.
Replication-based approaches initialize multiple FFNs from the same dense FFN and then learn separate weight values.
Some of them partially reinitialize these copies to encourage their differentiation~\citep{nakamura2025dropupcycling}.
Other methods construct smaller experts by partitioning or pruning pretrained FFNs~\citep{zhang-etal-2022-moefication,zhu-etal-2024-llama,feng-etal-2025-dive}, or introduce expert-specific adapters and low-rank updates ~\citep{wu-etal-2024-parameter,chen2024llavamolesparsemixturelora}.

We investigate a different parameterization: \emph{can experts be learned by selecting connections in a pretrained FFN?}
Learning a mask can change a network's function without changing its retained weights, as demonstrated by supermask learning and learnable LLM pruning~\citep{Ramanujan_2020_CVPR,NEURIPS2024_0e9a05f5,zhang2026boostinglargelanguagemodels,zhang2025rethinkingfinetuningunlockinghidden}.
In an MoE, several such masks can coexist over the same weights.
The router then chooses among alternative sparse computations.
This replaces expert-specific weight adaptation with expert-specific connectivity learning.

We propose \textbf{MASKerade}, which jointly learns a bank of binary expert masks and a token-level router over a frozen FFN.
For a given token, the router selects a small subset of masks, executes the corresponding masked FFNs, and combines their outputs using routing weights.
Conceptually, this resembles a \textbf{MASKerade (\emph{masquerade}, with ``qu'' replaced by ``k'')} in which tokens enter the same shared FFN but \emph{wear} different masks.
MASKerade studies a specific combination that consists of direct masking of the \emph{frozen pretrained FFN weights}, token-level sparse routing, and hard connectivity constraints.
We evaluate MASKerade through multimodal instruction tuning of
Qwen and Gemma backbones~\citep{yang2025qwen3technicalreport,gemmateam2025gemma3technicalreport}.
MASKerade attains the best performance against the external methods across five benchmarks.
Our contributions are:
\begin{itemize}
    \item We introduce MASKerade, a dense-to-MoE upcycling method that constructs experts as learned binary masks over frozen pretrained FFNs.
    \item MASKerade jointly optimizes expert masks and token routing under prescribed sparsity constraints, accommodating neuron-structured, semi-structured, and unstructured masks.
    \item Our experiments and analysis demonstrate that effective dense-to-MoE adaptation can be trained without updating the underlying pretrained FFN weights.
\end{itemize}

\section{Related Work}
\label{sec:related-work}

\noindent{\textbf{Sparse MoE and dense-to-MoE conversion.}}
Sparsely gated MoE layers use input-dependent expert selection to increase the available model capacity while restricting active computation~\citep{shazeer2017outrageouslylargeneuralnetworks,fedus2022switchtransformersscalingtrillion}.
At the routing level, MaskMoE uses fixed, token-frequency-dependent masks on routing logits to restrict the experts accessible to each vocabulary token~\citep{su2024maskmoeboostingtokenlevellearning}.
These masks constrain expert visibility rather than define sparse FFN subnetworks.
Sparse Upcycling initializes expert layers from pretrained dense checkpoints~\citep{komatsuzaki2023sparse}, and Drop-Upcycling uses partial reinitialization to reduce the similarity of copied experts~\citep{nakamura2025dropupcycling}.
MoEfication interprets FFN neurons as groups of experts and learns to select relevant groups~\citep{zhang-etal-2022-moefication}.
LLaMA-MoE combines expert partitioning with continual pretraining~\citep{zhu-etal-2024-llama}, while DIVE uses domain affinity
and calibration-dependent pruning to construct diverse experts~\citep{feng-etal-2025-dive}.
ToMoE learns dynamic structural pruning through a hypernetwork to convert dense models into routed, narrower computations~\citep{gao2026tomoe}.
MASKerade also starts from a dense model, but learns masks directly on its frozen FFN tensors instead of requiring independently fine-tuned FFN copies.


\noindent{\textbf{Learning sparse subnetworks.}}
Supermask methods learn useful subnetworks while keeping their underlying weight values fixed~\citep{Ramanujan_2020_CVPR}. For pretrained LLMs, Wanda selects weights using weight magnitudes and activation statistics without gradient-based mask training~\citep{sun2024simpleeffectivepruningapproach}.
SparseLLM formulates global pruning as coordinated optimization subproblems, using auxiliary variables to make the problem tractable~\citep{NEURIPS2024_522134ee}.
MaskLLM learns a distribution over legal semi-structured patterns using Gumbel-Softmax~\citep{NEURIPS2024_0e9a05f5}.
Beyond compression, Mask Fine-Tuning (MFT) uses the original fine-tuning objective to learn binary weight masks over a frozen, already fine-tuned LLM, seeking further capability improvements~\citep{zhang2026boostinglargelanguagemodels}.
\citet{zhang2025rethinkingfinetuningunlockinghidden} extend this approach to pretrained vision-language models, learning hard or soft masks over language-model and projector weights.
These methods motivate learning connectivity rather than only pruning by a fixed importance criterion.
MASKerade differs from a single learned subnetwork by maintaining multiple masks and selecting them conditionally for each token.

\noindent{\textbf{Mask-based and interpolated experts.}}
MoM explicitly studies learnable masks as an alternative to conventional experts~\citep{liu-etal-2025-masks}.
SMAT learns sparse interpolated experts for few-shot meta-tuning of vision models~\citep{chen2024unleashingpowermetatuningfewshot}.
SIMoE applies sparse expert discovery to LLM instruction tuning, where structured masks act on a shared trainable parameter delta added to the frozen pretrained parameters. Their main routing configuration merges experts at the instance level~\citep{chen-etal-2025-automatic}.
CDSP-MoE learns expert topology within a shared parameter subspace using gradient-conflict signals~\citep{gan2026mixtureofexpertsgradientconflictdrivensubspace}.
Our distinction is to route tokens through hard-masked versions of the pretrained FFN itself, with no trainable shared delta.

\section{Method}
\label{sec:method}

\subsection{Experts as Frozen-Weight Subnetworks}
\label{sec:mask-experts}

\begin{wrapfigure}{r}{0.5\textwidth}
    \centering
    \vspace{-12pt}
    \includegraphics[width=\linewidth]{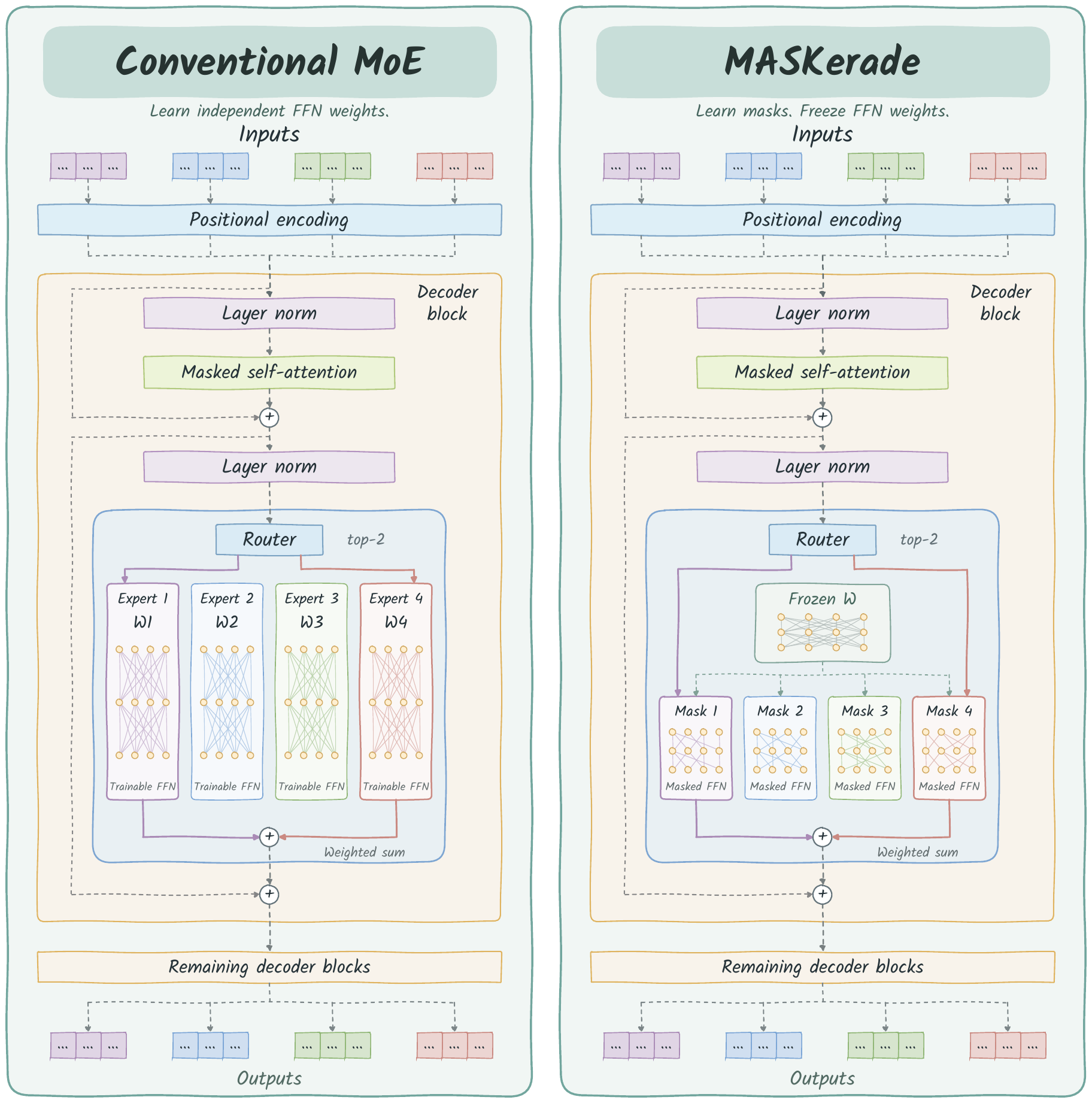}
    \caption{\textbf{Conventional MoE and MASKerade.} Tokens select independent FFNs (left) or learned masks over one frozen FFN (right).}
    \label{fig:arch}
    \vspace{-30pt}
\end{wrapfigure}


Let $\theta_0$ denote a pretrained dense model and $\mathcal{I}$ the set of FFN
layers to convert. In a converted layer $\ell$, write its frozen FFN parameters
as $W_\ell$. Conventional replicated experts learn separate parameter sets
$\{W_{\ell,e}\}_{e=1}^{E}$ initialized from $W_\ell$. MASKerade instead defines
\begin{equation}
    \begin{aligned}
    \widetilde W_{\ell,e}^{(a)}
    &= W_\ell^{(a)} \odot M_{\ell,e}^{(a)}, \\
    M_{\ell,e}^{(a)}
    &\in \{0,1\}^{\operatorname{shape}(W_\ell^{(a)})}.
    \end{aligned}
    \label{eq:masked-weights}
\end{equation}
where $a$ indexes the FFN projections and $\odot$ is elementwise multiplication.

For the gated FFNs used by our backbones, with gate, up, and down projections
$W^{(g)},W^{(u)}\in\mathbb{R}^{d_{\mathrm{ff}}\times d}$ and
$W^{(d)}\in\mathbb{R}^{d\times d_{\mathrm{ff}}}$, expert $e$ computes
\begin{equation}
    F_{\ell,e}(h)
    = \widetilde W_{\ell,e}^{(d)}
      \left[
        \phi_\ell\!\left(\widetilde W_{\ell,e}^{(g)}h\right)
        \odot
        \left(\widetilde W_{\ell,e}^{(u)}h\right)
      \right].
    \label{eq:expert-forward}
\end{equation}
In our configuration, the pretrained parameters remain frozen throughout
upcycling. 

\subsection{Token-Level Routing}
\label{sec:routing}

Routing selects from a learned bank of expert masks shared across tokens. A trainable linear router $R_\ell\in\mathbb{R}^{E\times d}$ maps each token's
hidden state to probabilities
$p_{\ell,e}(h)=\operatorname{softmax}(R_\ell h)_e$.
For $k$ experts, deterministic top-$k$ selection gives
$\mathcal{S}_\ell(h)=\operatorname{TopK}(p_\ell(h),k)$.
Our routing formulation uses deterministic top-$k$ expert selection
and dropless dispatch during both training and inference. 
The selected expert outputs are combined as
\begin{equation}
F_\ell^{\mathrm{MoE}}(h)
=
\sum_{e\in\mathcal{S}_\ell(h)}
g_{\ell,e}(h) F_{\ell,e}(h),
\qquad
g_{\ell,e}(h)
=
\begin{cases}
p_{\ell,e}(h), & k=1, \\[4pt]
\displaystyle
\frac{p_{\ell,e}(h)}
{\sum_{j\in\mathcal{S}_\ell(h)}p_{\ell,j}(h)}, & k\geq 2.
\end{cases}
\label{eq:routed-output}
\end{equation}
For $k\geq 2$, the coefficients sum to one over the selected experts.
For top-1 routing with $E>1$, retaining the selected expert's softmax
probability preserves a differentiable path from the task loss to the
router.
When $E=1$, routing will be bypassed.

\begin{table}[t]
\caption{Main comparison with existing methods on small models. All methods start from the same Zero-shot checkpoint. We learn masks as experts over frozen FFN weights. Best and second-best results in each metric are shown in \textbf{bold} and \underline{underlined}, respectively.
}
\label{tab:main}
\begin{center}
\small
\setlength{\tabcolsep}{3.5pt}
\resizebox{\textwidth}{!}{%
\begin{tabular}{clcccccc}
\toprule
& Method & Act. Params. & GQA {↑} & MME {↑} & POPE {↑} & SQA {↑} & TextVQA {↑} \\
\midrule

\multirow{13}{*}{\rotatebox{90}{Qwen3-1.7B}}
& Zero-shot~\citep{yang2025qwen3technicalreport}
& 1.72B
& 25.97 
& 962.0 
& 83.59 
& 60.93 
& 32.38 \\

& Dense MLP
& 1.72B
& \ms{58.41}{0.35}
& \ms{1300.4}{18.4}
& \ms{85.01}{0.24}
& \ms{61.16}{0.32}
& \ms{49.19}{0.32} \\

\cmidrule(l){2-8}

& MoEfication~\citep{zhang-etal-2022-moefication}
& 1.46B
& \ms{53.60}{0.20}
& \ms{1280.5}{15.8}
& \ms{84.75}{0.33}
& \ms{60.11}{0.96}
& \ms{49.94}{0.55} \\

& LLaVA-MoLE~\citep{chen2024llavamolesparsemixturelora}
& 1.91B
& \ms{58.05}{0.38}
& \ms{1295.0}{18.3}
& \ms{84.97}{0.36}
& \ms{60.24}{0.09}
& \ms{50.25}{0.06} \\

& PESC~\citep{wu-etal-2024-parameter} 
& 1.73B
& \ms{56.81}{0.36}
& \ms{1309.1}{11.3}
& \ms{84.31}{0.29}
& \ms{61.60}{0.57}
& \ms{49.95}{0.25} \\

& LLaMA-MoE~\citep{zhu-etal-2024-llama}
& 1.46B
& \ms{57.85}{0.16}
& \ms{1233.3}{17.5}
& \ms{85.45}{0.39}
& \ms{60.56}{0.58}
& \ms{50.90}{0.58} \\

& Drop-Upcycling~\citep{nakamura2025dropupcycling}
& 2.25B
& \underline{\ms{60.77}{0.24}}
& \ms{1312.9}{13.9}
& \ms{85.66}{0.19}
& \ms{61.60}{0.65}
& \ms{50.82}{0.07} \\

& MoM~\citep{liu-etal-2025-masks}
& 2.25B
& \ms{53.48}{0.13}
& \ms{1283.2}{16.0}
& \underline{\ms{85.73}{0.25}}
& \underline{\ms{62.37}{0.75}}
& \ms{50.49}{0.27} \\

& DIVE~\citep{feng-etal-2025-dive}
& 1.46B
& \ms{57.87}{0.16}
& \ms{1300.6}{16.7}
& \ms{83.80}{0.34}
& \ms{60.84}{0.76}
& \ms{50.55}{0.62} \\

& MoE-LLaVA~\citep{11353922}
& 2.25B
& \ms{60.06}{0.19}
& \underline{\ms{1315.3}{11.8}}
& \ms{85.19}{0.22}
& \ms{62.07}{0.76}
& \underline{\ms{51.36}{0.32}} \\

& ToMoE~\citep{gao2026tomoe}
& 1.46B
& \ms{58.54}{0.22}
& \ms{1302.2}{16.2} 
& \ms{85.66}{0.40} 
& \ms{61.61}{0.50} 
& \ms{50.77}{0.22} \\

\cmidrule(l){2-8}

& \textbf{[Ours]} MASKerade (Semi-structured 2:4)
& 1.72B
& \textbf{\ms{62.00}{0.28}}
& \textbf{\ms{1332.4}{11.9}}
& \textbf{\ms{86.49}{0.16}}
& \textbf{\ms{63.48}{0.57}}
& \textbf{\ms{52.50}{0.37}} \\

\midrule

\multirow{13}{*}{\rotatebox{90}{Gemma3-1B}}
& Zero-shot~\citep{gemmateam2025gemma3technicalreport}
& 1.00B
& 28.25
& 823.0
& 82.13
& 36.74
& 16.66 \\

& Dense MLP
& 1.00B
& \ms{56.32}{0.12}
& \ms{1233.4}{13.9}
& \ms{85.71}{0.16}
& \ms{52.20}{0.69}
& \ms{41.30}{0.31} \\

\cmidrule(l){2-8}

& MoEfication~\citep{zhang-etal-2022-moefication}
& 0.84B
& \ms{55.36}{0.28}
& \ms{1212.9}{13.5}
& \ms{80.92}{0.35}
& \ms{50.33}{0.55}
& \ms{40.68}{0.58} \\

& LLaVA-MoLE~\citep{chen2024llavamolesparsemixturelora}
& 1.11B
& \ms{55.66}{0.10}
& \ms{1244.8}{16.4}
& \ms{85.77}{0.24}
& \ms{51.56}{0.44}
& \ms{40.54}{0.83} \\

& PESC~\citep{wu-etal-2024-parameter}
& 1.00B
& \ms{57.54}{0.29}
& \ms{1267.1}{18.1}
& \ms{85.00}{0.36}
& \ms{53.91}{0.52}
& \ms{42.49}{0.71} \\

& LLaMA-MoE~\citep{zhu-etal-2024-llama}
& 0.84B
& \ms{57.67}{0.33}
& \ms{1234.7}{11.2}
& \ms{85.00}{0.42}
& \ms{52.71}{0.63}
& \ms{40.96}{0.67} \\

& Drop-Upcycling~\citep{nakamura2025dropupcycling}
& 1.31B
& \ms{58.94}{0.19}
& \ms{1287.5}{14.4}
& \ms{85.59}{0.40}
& \ms{54.24}{0.40}
& \underline{\ms{42.55}{0.68}} \\

& MoM~\citep{liu-etal-2025-masks}
& 1.31B
& \ms{56.79}{0.29}
& \ms{1249.1}{18.9}
& \ms{82.10}{0.40}
& \ms{50.07}{0.51}
& \ms{39.45}{0.79} \\

& DIVE~\citep{feng-etal-2025-dive}
& 0.85B
& \ms{51.66}{0.27}
& \ms{1129.3}{13.2}
& \ms{82.80}{0.28}
& \ms{50.84}{0.63}
& \ms{41.79}{0.58} \\

& MoE-LLaVA~\citep{11353922}
& 1.31B
& \underline{\ms{59.34}{0.45}}
& \underline{\ms{1288.0}{18.4}}
& \underline{\ms{85.80}{0.18}}
& \underline{\ms{54.83}{0.42}}
& \ms{41.91}{0.50} \\

& ToMoE~\citep{gao2026tomoe}
& 0.84B
& \ms{53.96}{0.15} 
& \ms{1123.0}{17.9} 
& \ms{81.37}{0.43} 
& \ms{51.98}{0.52} 
& \ms{40.63}{0.32} \\

\cmidrule(l){2-8}

& \textbf{[Ours]} MASKerade (Semi-structured 2:4)
& 1.00B
& \textbf{\ms{60.07}{0.27}}
& \textbf{\ms{1293.8}{11.9}}
& \textbf{\ms{86.41}{0.10}}
& \textbf{\ms{55.30}{0.47}}
& \textbf{\ms{42.90}{0.47}} \\

\bottomrule
\end{tabular}}
\end{center}
\vspace{-16pt}
\end{table}

\subsection{Mask Parameterization and Granularity}
\label{sec:mask-granularity}

Each mask is obtained from continuous trainable scores $S$ using a hard
selection operator $\mathcal{H}$, so $M=\mathcal{H}(S)$ is binary. We denote the retained fraction by $\rho$ and consider three families.

\noindent{\textbf{Semi-structured.}}
For an $N{:}M$ pattern, each row of a weight matrix is partitioned into
contiguous groups of $M$ entries along its dimension.
For each group $B$ we keep the $N$ largest score magnitudes:
\begin{equation}
    M_{\ell,e,i}^{(a)}
      = \mathbf{1}\!\left[
        i\in\operatorname{TopN}_{j\in B}
                   \left|S_{\ell,e,j}^{(a)}\right|
        \right], \qquad
    \sum_{i\in B}M_{\ell,e,i}^{(a)}=N .
    \label{eq:nm-mask}
\end{equation}

\noindent{\textbf{Unstructured.}}
For each expert's projection tensor, let $P$ denote its number of entries
and $q=\lceil\rho P\rceil$ the retained count. We select the $q$ largest
score magnitudes over all flattened positions:
\[
    M_{\ell,e,i}^{(a)}
      = \mathbf{1}\!\left[
        i\in\operatorname{TopK}_{1\leq j\leq P}
             \left(\left|S_{\ell,e,j}^{(a)}\right|,q\right)
        \right], \qquad
    \sum_{i=1}^{P}M_{\ell,e,i}^{(a)}=q .
\]

\noindent{\textbf{Structured.}}
Each expert instead learns a score vector over the $d_{\mathrm{ff}}$ intermediate
neurons, retaining a fraction $\rho$. Its binary mask $m_{\ell,e}$ is shared
across the corresponding gate/up rows and down columns:
\begin{equation}
    M_{\ell,e}^{(g)}=M_{\ell,e}^{(u)}
         =m_{\ell,e}\mathbf{1}_{d}^{\top}, \qquad
    M_{\ell,e}^{(d)}=\mathbf{1}_{d}m_{\ell,e}^{\top}.
    \label{eq:neuron-mask}
\end{equation}

\subsection{Joint Mask and Router Training}
\label{sec:optimization}

\noindent{\textbf{Score initialization.}}
We consider two initialization schemes for the mask scores.
With \emph{random initialization}, each expert's scores are sampled
independently using Kaiming-uniform initialization.
\begin{wrapfigure}{r}{0.35\textwidth}
    \centering
    \includegraphics[width=\linewidth]{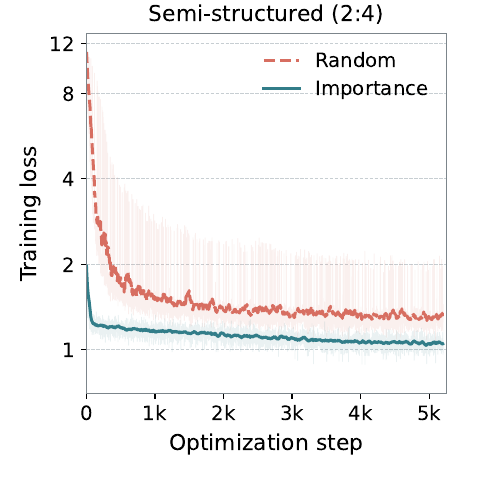}
    \vspace{-20pt}
    \caption{Initialization loss for 2:4 masks on Qwen3-1.7B.}
    \label{fig:mask-initialization}
    \vspace{-40pt}
\end{wrapfigure}
With \emph{importance initialization}, weight-level scores instead use
the pretrained weight magnitudes with independent expert-specific noise:
\begin{equation}
    \begin{aligned}
        S_{\ell,e}^{(a)}
        &= |W_\ell^{(a)}| + \epsilon_{\ell,e}^{(a)}, \\
        \epsilon_{\ell,e,i}^{(a)}
        &\sim \mathcal{N}\!\left(
        0,
        \left[
        \alpha_{\mathrm{init}}\,
        \operatorname{std}(|W_\ell^{(a)}|)
        \right]^2
        \right).
        \end{aligned}
    \label{eq:initialization}
\end{equation}
$\alpha_{\mathrm{init}}\geq 0$ controls the relative noise scale.
The magnitude term favors high-magnitude connections, while independent
perturbations help break symmetry between expert initializations.
For neuron-structured masks, the importance of neuron $j$ is
$u_{\ell,j}=(\operatorname{mean}|W_{\ell,j,:}^{(g)}|
+\operatorname{mean}|W_{\ell,j,:}^{(u)}|
+\operatorname{mean}|W_{\ell,:,j}^{(d)}|)/3$.
We initialize its score as $S_{\ell,e,j}=u_{\ell,j}+\epsilon_{\ell,e,j}$,
with independent Gaussian noise of standard deviation
$\alpha_{\mathrm{init}}\operatorname{std}(u_\ell)$ across neurons and experts.
Thus, structured importance aggregates the three weight slices affected
by the same neuron mask.

The loss curve comparison in Fig.~\ref{fig:mask-initialization}
shows a substantially lower initial loss with importance initialization,
as well as a lower training loss near the end of the run.
Motivated by this observation, we adopt importance initialization as the
default for subsequent MASKerade experiments. Tab.~\ref{tab:add-init} and Fig.~\ref{fig:mask-initialization-all} show more analysis.

\begin{table}[t]
\caption{Main comparison with existing methods on large models. All methods start from the same Zero-shot checkpoint. We learn masks as experts over frozen FFN weights. Best and second-best results in each metric are shown in \textbf{bold} and \underline{underlined}, respectively.
}
\label{tab:main-large}
\begin{center}
\small
\setlength{\tabcolsep}{3.5pt}
\resizebox{\textwidth}{!}{%
\begin{tabular}{clcccccc}
\toprule
& Method & Act. Params. & GQA {↑} & MME {↑} & POPE {↑} & SQA {↑} & TextVQA {↑} \\
\midrule

\multirow{13}{*}{\rotatebox{90}{Qwen3.6-27B}}
& Zero-shot~\citep{yang2025qwen3technicalreport}
& 26.90B
& 61.39 
& 1692.1 
& 86.23 
& 92.74 
& 71.65
\\

& Dense MLP
& 26.90B
& 64.13 
& 1716.0 
& 86.90 
& 93.38 
& 73.41
\\

\cmidrule(l){2-8}

& MoEfication~\citep{zhang-etal-2022-moefication}
& 22.62B
& 61.05 
& 1691.2 
& 86.15 
& 91.55 
& 72.00
\\

& LLaVA-MoLE~\citep{chen2024llavamolesparsemixturelora}
& 29.82B
& 65.71 
& 1702.5 
& 86.47 
& 92.17 
& 74.08
\\

& PESC~\citep{wu-etal-2024-parameter}
& 26.94B
& 65.61 
& \underline{1717.4}
& 87.08 
& 91.18 
& 77.33
\\

& LLaMA-MoE~\citep{zhu-etal-2024-llama}
& 22.62B
& 64.33 
& 1715.5 
& 86.78 
& 93.46 
& 78.63
\\

& Drop-Upcycling~\citep{nakamura2025dropupcycling}
& 35.45B
& \underline{65.75}
& 1715.3 
& 87.26 
& 94.12 
& 79.04
\\

& MoM~\citep{liu-etal-2025-masks}
& 35.45B 
& 61.27 
& 1689.3 
& 86.29 
& 88.82 
& 79.95 \\

& DIVE~\citep{feng-etal-2025-dive}
& 22.65B
& 64.58 
& 1708.0 
& 87.16 
& 90.83 
& 79.68
\\

& MoE-LLaVA~\citep{11353922}
& 35.45B
& 65.58 
& 1715.4 
& 87.47 
& \underline{94.42}
& \underline{80.27}
\\

& ToMoE~\citep{gao2026tomoe}
& 22.62B 
& 61.87 
& 1681.2 
& \underline{87.58}
& 89.57 
& 71.80
\\

\cmidrule(l){2-8}

& \textbf{[Ours]} MASKerade (Semi-structured 2:4)
& 26.90B
& \textbf{66.37}
& \textbf{1724.9}
& \textbf{87.68}
& \textbf{94.79}
& \textbf{81.08}
\\

\midrule

\multirow{13}{*}{\rotatebox{90}{Gemma4-31B}}
& Zero-shot~\citep{gemmateam2026gemma4technicalreport}
& 30.70B
& 47.46 
& 1573.8 
& 82.48 
& 83.24 
& 68.28
\\

& Dense MLP
& 30.70B
& 60.81 
& 1626.6 
& 86.00 
& 83.59 
& 72.67
\\

\cmidrule(l){2-8}

& MoEfication~\citep{zhang-etal-2022-moefication}
& 25.50B
& 58.59 
& 1559.6 
& 79.74 
& 81.17 
& 68.02
\\

& LLaVA-MoLE~\citep{chen2024llavamolesparsemixturelora}
& 34.24B
& 61.24 
& 1578.3 
& 86.52 
& 74.96 
& 73.26
\\

& PESC~\citep{wu-etal-2024-parameter}
& 30.74B
& 63.11 
& 1651.3 
& \underline{87.07}
& \underline{86.07}
& 74.96
\\

& LLaMA-MoE~\citep{zhu-etal-2024-llama}
& 25.50B
& 63.44
& 1629.5
& 86.39
& 77.24
& 74.52
\\

& Drop-Upcycling~\citep{nakamura2025dropupcycling}
& 41.10B
& \underline{65.22}
& 1676.2 
& 86.86 
& 81.51 
& \underline{75.11}
\\

& MoM~\citep{liu-etal-2025-masks}
& 41.10B
& 58.81 
& 1581.4 
& 82.89 
& 84.28 
& 68.87
\\

& DIVE~\citep{feng-etal-2025-dive}
& 25.53B
& 61.25
& 1573.3
& 83.36
& 73.82
& 72.60
\\

& MoE-LLaVA~\citep{11353922}
& 41.10B
& 64.86 
& \underline{1678.4}
& 86.63 
& 84.50 
& 74.88
\\

& ToMoE~\citep{gao2026tomoe}
& 25.50B 
& 60.58 
& 1575.6 
& 85.34 
& 83.29 
& 69.37
\\

\cmidrule(l){2-8}

& \textbf{[Ours]} MASKerade (Semi-structured 2:4)
& 30.70B
& \textbf{65.30}
& \textbf{1694.3}
& \textbf{87.33}
& \textbf{87.62}
& \textbf{75.44}
\\

\bottomrule
\end{tabular}}
\end{center}
\vspace{-16pt}
\end{table}

\noindent{\textbf{Gradient estimator.}}
Hard selection is not differentiable. Following straight-through and
supermask learning~\citep{bengio2013estimatingpropagatinggradientsstochastic,zhang2026boostinglargelanguagemodels}, we use
hard masks in the forward pass and a straight-through surrogate to
propagate gradients to the continuous mask scores in the backward pass.
For each masked projection, the learning signal at the mask is
\begin{equation}
    \frac{\partial\mathcal{L}}{\partial M_{\ell,e}^{(a)}}
      =
    W_\ell^{(a)}\odot
    \frac{\partial\mathcal{L}}
         {\partial\widetilde W_{\ell,e}^{(a)}} .
    \label{eq:ste}
\end{equation}
For neuron-structured masks, gradients from the incident
connections are accumulated into the shared neuron scores. All masks are
recomputed from the updated scores.

\noindent{\textbf{Training objective.}}
Mask scores and routers are learned jointly through
supervised autoregressive loss and a load-balancing term:
\begin{equation}
    \min_{\{S_\ell,R_\ell\}_{\ell\in\mathcal{I}}}
    \mathcal{L}_{\mathrm{SFT}}(\theta_0,S,R)
       + \lambda_{\mathrm{bal}}
         \sum_{\ell\in\mathcal{I}}\mathcal{L}_{\mathrm{bal}}^{(\ell)},
    \qquad \theta_0\ \text{fixed},
    \label{eq:objective}
\end{equation}
where $\mathcal{L}_{\mathrm{SFT}}$ is the mean negative log-likelihood of the
supervised response tokens, and $\lambda_{\mathrm{bal}}$ controls the
strength of load balancing. Following the probability-load balancing
principle of~\citep{fedus2022switchtransformersscalingtrillion}, we measure expert usage over all
selected assignments and define the penalty over $T$ tokens entering a
converted layer as
\begin{equation}
    \bar p_{\ell,e}=\frac{1}{T}\sum_{t=1}^{T}p_{\ell,e}(h_t),\quad
    f_{\ell,e}=\frac{1}{T}\sum_{t=1}^{T}
       \mathbf{1}\!\left[e\in\mathcal{S}_\ell(h_t)\right],\quad
    \mathcal{L}_{\mathrm{bal}}^{(\ell)}
       = E\sum_{e=1}^{E}\bar p_{\ell,e}f_{\ell,e}.
    \label{eq:balance}
\end{equation}
$\bar p_{\ell,e}$ measures the mean routing probability, and
$f_{\ell,e}$ measures the fraction of tokens assigned to expert $e$.
Each token contributes to all $k$ selected experts, so
$\sum_{e=1}^{E}f_{\ell,e}=k$.

\begin{table}[t]
\caption{Mask-based MoE comparison. Best and second-best results in each metric are shown in \textbf{bold} and \underline{underlined}, respectively.}
\label{tab:mask-baselines}
\begin{center}
\small
\setlength{\tabcolsep}{3.5pt}
\resizebox{\textwidth}{!}{%
\begin{tabular}{clccccc}
\toprule
& Method & GQA & MME & POPE & SQA & TextVQA \\
\midrule

\multirow{10}{*}{\rotatebox{90}{Qwen3-1.7B}}
& Zero-shot
& 25.97 
& 962.0 
& 83.59 
& 60.93 
& 32.38 \\

& Dense MLP
& \ms{58.41}{0.35}
& \ms{1300.4}{18.4}
& \ms{85.01}{0.24}
& \ms{61.16}{0.32}
& \ms{49.19}{0.32} \\

\cmidrule(l){2-7}

& Random 2:4
& \ms{43.36}{0.30} 
& \ms{871.8}{21.3} 
& \ms{82.15}{0.16} 
& \ms{40.75}{0.42}
& \ms{28.22}{0.48} \\

& Magnitude 2:4
& \ms{51.54}{0.26} 
& \ms{1127.6}{14.7} 
& \ms{85.38}{0.53} 
& \ms{58.34}{0.80} 
& \ms{43.15}{0.71} \\

& Wanda 2:4~\citep{sun2024simpleeffectivepruningapproach}
& \ms{51.97}{0.53} 
& \ms{1148.2}{23.8} 
& \ms{83.96}{0.44} 
& \ms{60.81}{0.64} 
& \ms{44.01}{0.27} \\

& Wanda Unstructured~\citep{sun2024simpleeffectivepruningapproach}
& \ms{52.24}{0.48} 
& \ms{1187.4}{37.4} 
& \ms{86.06}{0.27} 
& \ms{63.01}{0.52} 
& \ms{45.21}{0.22} \\

\cmidrule(l){2-7}

& \textbf{[Ours]} Structured
& \ms{58.81}{0.12} 
& \ms{1302.0}{18.3} 
& \ms{85.06}{0.48} 
& \ms{61.58}{0.52} 
& \ms{51.05}{0.24} \\

& \textbf{[Ours]} Unstructured
& \textbf{\ms{62.06}{0.30}} 
& \underline{\ms{1315.8}{11.2}}
& \underline{\ms{86.27}{0.32}} 
& \textbf{\ms{63.71}{0.42}} 
& \textbf{\ms{52.83}{0.13}} \\

& \textbf{[Ours]} Semi-tructured 2:4
& \underline{\ms{62.00}{0.28}}
& \textbf{\ms{1332.4}{11.9}}
& \textbf{\ms{86.49}{0.16}}
& \underline{\ms{63.48}{0.57}}
& \underline{\ms{52.50}{0.37}} \\

\midrule

\multirow{10}{*}{\rotatebox{90}{Gemma3-1B}}
& Zero-shot
& 28.25
& 823.0
& 82.13
& 36.74
& 16.66 \\

& Dense MLP
& \ms{56.32}{0.12}
& \ms{1233.4}{13.9}
& \ms{85.71}{0.16}
& \ms{52.20}{0.69}
& \ms{41.30}{0.31} \\

\cmidrule(l){2-7}

& Random 2:4
& \ms{36.45}{0.64} 
& \ms{732.8}{20.6} 
& \ms{64.37}{0.69} 
& \ms{23.24}{0.82} 
& \ms{13.50}{0.78} \\

& Magnitude 2:4
& \ms{46.10}{0.36} 
& \ms{920.5}{18.3} 
& \ms{83.10}{0.56} 
& \ms{34.03}{0.52} 
& \ms{28.87}{0.58} \\

& Wanda 2:4~\citep{sun2024simpleeffectivepruningapproach}
& \ms{47.45}{0.16}
& \ms{904.8}{6.4}
& \ms{82.87}{0.70}
& \ms{36.23}{0.62}
& \ms{30.25}{0.75} \\

& Wanda Unstructured~\citep{sun2024simpleeffectivepruningapproach}
& \ms{47.38}{0.17} 
& \ms{1001.0}{15.7} 
& \ms{80.62}{0.78} 
& \ms{35.94}{0.43} 
& \ms{32.34}{0.97} \\

\cmidrule(l){2-7}

& \textbf{[Ours]} Structured
& \ms{56.88}{0.47} 
& \ms{1253.5}{12.9} 
& \ms{85.81}{0.34} 
& \ms{52.33}{0.84} 
& \ms{41.58}{0.37} \\

& \textbf{[Ours]} Unstructured
& \textbf{\ms{60.38}{0.11}} 
& \underline{\ms{1288.9}{7.1}}
& \underline{\ms{85.87}{0.58}}
& \textbf{\ms{56.68}{0.74}} 
& \underline{\ms{42.41}{0.49}} \\

& \textbf{[Ours]} Semi-tructured 2:4
& \underline{\ms{60.07}{0.27}}
& \textbf{\ms{1293.8}{11.9}}
& \textbf{\ms{86.41}{0.10}}
& \underline{\ms{55.30}{0.47}}
& \textbf{\ms{42.90}{0.47}} \\

\bottomrule
\end{tabular}}
\end{center}
\end{table}

\section{Experiments}

We evaluate MASKerade as a dense-to-MoE adaptation method for multimodal
instruction tuning.

\subsection{Experimental Setup}
\label{sec:experimental-setup}

\noindent{\textbf{Backbones and training data.}}
We conduct our experiments on both small and large Vision-Language Models (VLMs). For small models, we use LLaVA-style vision-language models~\citep{NEURIPS2023_6dcf277e}
with Qwen3-1.7B~\citep{yang2025qwen3technicalreport} and
Gemma3-1B~\citep{gemmateam2025gemma3technicalreport} language backbones.
Both use a CLIP ViT-L/14 image encoder~\citep{radford2021learningtransferablevisualmodels}.
For large models, we use Qwen3.6-27B~\citep{yang2025qwen3technicalreport} and
Gemma4-31B~\citep{gemmateam2026gemma4technicalreport} as the VLM backbones.
Adaptation uses the 665K instruction mixture in the
MoE-LLaVA pipeline~\citep{11353922}.

\noindent{\textbf{Baselines and benchmarks.}}
We compare with representative strategies.
We adapt the baseline expert constructions to a common starting VLM and
instruction dataset within each backbone.
Expert widths, trainable modules, and learning rates remain
method-specific.
We evaluate on five benchmarks, including GQA~\citep{hudson2018gqa},
MME~\citep{fu2023mme},
POPE~\citep{li-etal-2023-evaluating},
ScienceQA~\citep{lu2022learn},
and TextVQA~\citep{singh2019towards}.

\noindent{\textbf{Implementation details.}}
We convert alternating FFN layers, with 14 of 28 for Qwen3-1.7B and 13 of 26
for Gemma3-1B, and 32 of 64 for Qwen3.6-27B and 30 of 60 for Gemma4-31B (the comparison of different layers is in Tab.~\ref{tab:placement}).
Layer indices are zero-based and even.
Each converted layer has $E=4$ mask experts with top-$2$ routing.
We use a cosine learning-rate schedule with 3\% warmup, zero weight decay,
and bfloat16 training. The default initialization noise is
$\alpha_{\mathrm{init}}=0.1$ and the balance coefficient is
$\lambda_{\mathrm{bal}}=0.01$.
We run most experiments with three random seeds and report the mean and
sample standard deviation.
More implementation details are in Tab.~\ref{tab:maskerade-implementation}.

\subsection{Main Results}
\label{sec:main-results}

\noindent{\textbf{Main Comparison.}}
Tab.~\ref{tab:main} and Tab.~\ref{tab:main-large} compare MASKerade with representative MoE adaptation methods from the same \emph{Zero-shot} checkpoint for each
backbone. Among the results, MASKerade achieves the highest reported performance on all five benchmarks. 
\emph{Act. Params.} denotes nominal activated language-model parameter counts, with overlapping masked weights counted separately across selected experts,
rather than unique weights, exact FLOPs, or measured latency.
Its nominal activated-parameter counts match Dense MLP and are below most baselines. 
We also provide a low-budget MASKerade compared with the lowest activated-computation methods in Tab.~\ref{tab:compmatch}.
Overall, these results support MASKerade as an effective alternative to learning independent experts.

\begin{table}[t]
\caption{Router learning and inference-time interventions with four 2:4 experts and top-2 routing. Frozen-random routers remain fixed during mask learning.}
\label{tab:swap}
\begin{center}
\small
\begin{tabular}{lccccc}
\toprule
Setting & GQA {↑} & MME {↑} & POPE {↑} & SQA {↑} & TextVQA {↑} \\
\midrule

{\emph{Qwen3-1.7B}} \\

Trained
& \ms{62.00}{0.28}
& \ms{1332.4}{11.9}
& \ms{86.49}{0.16}
& \ms{63.48}{0.57}
& \ms{52.50}{0.37} \\

Random (frozen during training)
& \ms{60.02}{0.07} 
& \ms{1307.7}{21.0} 
& \ms{85.96}{0.45} 
& \ms{61.27}{1.05} 
& \ms{50.97}{0.28} \\

Random (after training)
& \ms{59.30}{0.39}
& \ms{1297.8}{12.6}
& \ms{84.60}{0.11}
& \ms{60.32}{0.66}
& \ms{49.71}{0.58} \\

Label-permuted
& \ms{56.86}{0.52} 
& \ms{1270.0}{12.2} 
& \ms{83.82}{0.66} 
& \ms{60.82}{0.84} 
& \ms{49.63}{0.61} \\


\midrule

{\emph{Gemma3-1B}} \\

Trained
& \ms{60.07}{0.27}
& \ms{1293.8}{11.9}
& \ms{86.41}{0.10}
& \ms{55.30}{0.47}
& \ms{42.90}{0.47} \\

Random (frozen during training)
& \ms{58.92}{0.14} 
& \ms{1267.1}{16.1} 
& \ms{85.37}{0.74} 
& \ms{52.67}{0.19} 
& \ms{41.36}{0.43} \\

Random (after training)
& \ms{57.89}{0.18}
& \ms{1272.1}{17.7}
& \ms{85.04}{0.63}
& \ms{51.26}{0.80}
& \ms{39.40}{0.55} \\

Label-permuted
& \ms{56.13}{0.28} 
& \ms{1226.9}{16.1} 
& \ms{84.99}{0.58} 
& \ms{46.75}{0.99} 
& \ms{40.29}{0.65} \\


\bottomrule
\end{tabular}
\end{center}
\end{table}

\begin{table}[t]
\centering
\renewcommand{\arraystretch}{0.95}

\begin{minipage}[t]{0.48\linewidth}
\centering
\captionof{table}{Mask-score initialization on Qwen3-1.7B (four experts, top-2, 50\% retention).}
\label{tab:add-init}

\scriptsize
\setlength{\tabcolsep}{2pt}
\begin{adjustbox}{max width=\linewidth}
\begin{tabular}{lccccc}
\toprule
Initialization & GQA {↑} & MME {↑} & POPE {↑} & SQA {↑} & TextVQA {↑} \\
\midrule

\multicolumn{6}{l}{\emph{Structured 50\%}} \\
Random & 58.51 & 1290.0 & 84.11 & 60.21 & 49.82 \\
Importance & 58.90 & 1312.0 & 84.56 & 62.08 & 50.85 \\
Importance-permuted & 51.54 & 1060.0 & 84.01 & 57.21 & 43.76 \\

\midrule

\multicolumn{6}{l}{\emph{Unstructured 50\%}} \\
Random & 60.20 & 1303.9 & 85.23 & 58.95 & 48.99 \\
Importance & 62.06 & 1322.4 & 86.27 & 63.71 & 52.94 \\
Importance-permuted & 60.88 & 1242.8 & 84.93 & 59.44 & 48.24 \\

\midrule

\multicolumn{6}{l}{\emph{Semi-structured 2:4}} \\
Random & 54.44 & 1101.1 & 84.65 & 55.43 & 39.47 \\
Importance & 62.08 & 1335.1 & 86.63 & 63.99 & 52.86 \\
Importance-permuted & 60.68 & 1280.5 & 85.41 & 55.23 & 47.46 \\

\bottomrule
\end{tabular}
\end{adjustbox}
\end{minipage}
\hfill
\begin{minipage}[t]{0.48\linewidth}
\centering
\captionof{table}{Per-expert retention across mask granularities on Qwen3-1.7B (four experts, top-2).}
\label{tab:ratio}

\scriptsize
\setlength{\tabcolsep}{2pt}
\begin{adjustbox}{max width=\linewidth}
\begin{tabular}{lccccc}
\toprule
Family & GQA {↑} & MME {↑} & POPE {↑} & SQA {↑} & TextVQA {↑} \\
\midrule

\multicolumn{6}{l}{\emph{Structured}} \\
25\% FFN parameters
& 57.00 
& 1299.8 
& 83.97 
& 59.19 
& 48.60 \\

50\% FFN parameters
& 58.90 
& 1312.0 
& 84.56 
& 62.08 
& 50.85 \\

75\% FFN parameters
& 60.99 
& 1315.4 
& 85.80 
& 62.57 
& 50.38 \\

\midrule

\multicolumn{6}{l}{\emph{Unstructured}} \\
25\% FFN parameters
& 61.70 
& 1322.0 
& 85.86 
& 60.69 
& 50.31 \\

50\% FFN parameters
& 62.06 
& 1322.4 
& 86.27 
& 63.71 
& 52.94 \\

75\% FFN parameters
& 62.13 
& 1375.7 
& 86.42 
& 65.88 
& 54.26 \\

\midrule

\multicolumn{6}{l}{\emph{Semi-structured}} \\
1:4 mask
& 61.20 
& 1312.2 
& 85.46 
& 61.29 
& 51.11 \\

2:4 mask
& 62.08 
& 1335.1 
& 86.63 
& 63.99 
& 52.86 \\

3:4 mask
& 62.92 
& 1338.9 
& 87.22 
& 65.94 
& 53.74 \\

\bottomrule
\end{tabular}
\end{adjustbox}
\end{minipage}

\end{table}

\noindent{\textbf{Mask construction and granularity.}}
Tab.~\ref{tab:mask-baselines} compares learned mask experts with random,
magnitude-based, and Wanda-based~\citep{sun2024simpleeffectivepruningapproach} mask construction. 
Fixed-mask experts obtain diversity through independent random initialization (Random), expert-specific score perturbations (Magnitude and Wanda
Unstructured), or source-specific calibration (Wanda 2:4).
Magnitude and Wanda
masks improve over random masks, but our learned 2:4 experts achieve
higher reported means than all fixed-mask baselines on all five
benchmarks. These comparisons support optimizing the
retained connections rather than relying solely on predetermined mask
construction.

Among the learned variants, both unstructured and semi-structured masks
outperform neuron-structured masks on all five benchmarks for both
backbones, and neither weight-level variant dominates the other.
Because NVIDIA GPUs support 2:4 sparsity for accelerated computation, we use the 2:4 mask as our primary configuration and report it throughout the main experiments.

\begin{table}[t]
\caption{Matched nominal activation budgets on Qwen3-1.7B. Calculation denotes retained density per expert multiplied by the number of activated experts.}
\label{tab:compmatch}
\begin{center}
\small
\resizebox{\textwidth}{!}{%
\begin{tabular}{lcccccccc}
\toprule
Configuration & Masks & Calculation & Act.\ params & GQA {↑} & MME {↑} & POPE {↑} & SQA {↑} & TextVQA {↑} \\
\midrule

MoEfication~\citep{zhang-etal-2022-moefication}
& -
& -
& 1.46B
& \ms{53.60}{0.20}
& \ms{1280.5}{15.8}
& \ms{84.75}{0.33}
& \ms{60.11}{0.96}
& \ms{49.94}{0.55} \\

LLaMA-MoE~\citep{zhu-etal-2024-llama}
& -
& -
& 1.46B
& \ms{57.85}{0.16}
& \ms{1233.3}{17.5}
& \ms{85.45}{0.39}
& \ms{60.56}{0.58}
& \ms{50.90}{0.58} \\

DIVE~\citep{feng-etal-2025-dive}
& -
& -
& 1.46B
& \ms{57.87}{0.16}
& \ms{1300.6}{16.7}
& \ms{83.80}{0.34}
& \ms{60.84}{0.76}
& \ms{50.55}{0.62} \\

ToMoE~\citep{gao2026tomoe}
& -
& -
& 1.46B
& \ms{58.54}{0.22}
& \ms{1302.2}{16.2} 
& \ms{85.66}{0.40} 
& \ms{61.61}{0.50} 
& \ms{50.77}{0.22} \\

\cmidrule{1-9}

1 mask only
& 2:4
& $50\%\times1$
& 1.46B
& \ms{56.51}{0.32} 
& \ms{1258.1}{15.5}
& \ms{84.50}{0.25}
& \ms{59.67}{0.56}
& \ms{49.13}{0.68} \\

[Ours] 4e/top-2 
& 1:4
& $25\%\times2$
& 1.46B 
& \ms{61.40}{0.32} 
& \ms{1322.2}{11.8} 
& \ms{85.96}{0.46} 
& \ms{61.79}{0.53} 
& \ms{50.94}{0.34} \\

[Ours] 4e/top-1
& 2:4
& $50\%\times1$
& 1.46B
& \ms{61.85}{0.31}
& \ms{1326.6}{4.8}
& \ms{86.29}{0.07}
& \ms{62.85}{0.32}
& \ms{51.76}{0.48} \\

[Ours] 4e/top-2
& 2:4
& $50\%\times2$
& 1.72B
& \ms{62.00}{0.28}
& \ms{1332.4}{11.9}
& \ms{86.49}{0.16}
& \ms{63.48}{0.57}
& \ms{52.50}{0.37} \\





\bottomrule
\end{tabular}}
\end{center}
\end{table}

\begin{table}[t]
\caption{Standalone FFN-block latency on an H100 (BF16, CUDA graphs, balanced synthetic routing). Ratios report relative latency.}
\label{tab:speed-shapes}
\centering
\small
\setlength{\tabcolsep}{4pt}

\begin{adjustbox}{max width=\linewidth}
\begin{tabular}{lcccccc}
\toprule
Model
& dense (ms)
& Vanilla (ms)
& ours (ms)
& Vanilla ($\times$dense)
& ours ($\times$dense)
& ours ($\times$Vanilla) \\
\midrule
Qwen3-1.7B   & 1.13 & 2.78  & 2.59  & 2.46 & 2.29 & 0.93 \\
Gemma3-1B    & 0.78 & 2.04  & 2.02  & 2.62 & 2.59 & 0.99 \\
Qwen3.6-27B  & 7.29 & 15.14 & 11.96 & 2.08 & 1.64 & 0.79 \\
Gemma-4-31B  & 9.62 & 20.69 & 15.52 & 2.15 & 1.61 & 0.75 \\
\bottomrule
\end{tabular}
\end{adjustbox}
\vspace{-4mm}
\end{table}

\subsection{Ablations and Analysis}
\label{sec:analysis}

\noindent{\textbf{Router optimization.}}
Tab.~\ref{tab:swap} distinguishes router optimization during training
from interventions applied after training. The frozen-random setting
keeps randomly initialized router parameters fixed while learning the
masks. Jointly learning masks and routers produces higher reported means on all five benchmarks. Replacing the trained router with a randomly initialized one at evaluation, while retaining the learned masks, reduces the model performance. The label-permuted intervention changes which learned mask each expert index selects, which also worsens model performance. This sensitivity to the index-to-mask correspondence supports the importance of the learned binding between the router and masks.

\noindent{\textbf{Effect of score initialization.}}
Tab.~\ref{tab:add-init} compares random, importance, and
importance-permuted initialization for learned masks across the three
granularities. Importance initialization achieves the highest scores on
all five benchmarks within each family. Its advantage over random
initialization is particularly pronounced for 2:4 masks. The importance-permuted control disrupts the alignment of the importance
prior with its original parameter locations. This supports retaining
location-specific importance information when initializing the scores.
Together, these observations support the default importance initialization across mask families.

\noindent{\textbf{Mask ratio.}}
Tab.~\ref{tab:ratio} varies the per-expert retained fraction from 25\%
to 75\% across the three mask families on Qwen3-1.7B, with four experts
and top-2 routing. The relative performance of mask families depends on retention. At 25\%, unstructured masks perform best on most benchmarks. Instead, at 50\% and 75\%, the best results are split between
unstructured and semi-structured masks.

\noindent{\textbf{Matched activation budgets.}}
Tab.~\ref{tab:compmatch} compares low-budget MASKerade variants with a
single learned mask and existing MoE methods at the lowest activated parameters on Qwen3-1.7B in the main table. The two MASKerade variants
evaluate either one 2:4 expert or two 1:4 experts per token, both using
half the nominal arithmetic of one dense pass through each converted
FFN. Both variants achieve higher reported means than the four external
baselines on all five benchmarks. Relative to a single learned 2:4 mask,
the 4e/top-1 configuration improves all five benchmarks at the same
nominal budget. At this budget, 2:4/top-1 also outperforms 1:4/top-2
on all five benchmarks. The default 2:4/top-2 configuration improves
further but increases nominal activated parameters from 1.46B to 1.72B.
These results compare complete routed configurations and do not isolate
a benefit from activating more experts alone.

\noindent{\textbf{Inference speed.}}
Tab.~\ref{tab:speed-shapes} compares a standalone SwiGLU microbenchmark
at the FFN dimensions of the four backbones. The dense reference
evaluates one FFN pass, while Vanilla and MASKerade evaluate
four-expert top-2 blocks with synthetic balanced expert assignments.
Measurements use bfloat16 on an H100 GPU under CUDA-graph replay.
The MASKerade's weight compression and algorithm selection occur before
timing. The timed block includes router operations, token dispatch,
expert computation, and weighted output combination. The normalized
columns report latency ratios. Relative to Vanilla, MASKerade takes $0.93\times$ and $0.99\times$ the
time at the Qwen3-1.7B and Gemma3-1B shapes, respectively.
The reduction is larger at the Qwen3.6-27B and Gemma4-31B shapes,
where the ratios are $0.79\times$ and $0.75\times$, corresponding to
approximately 21\% and 25\% lower latency. The benefit of MASKerade is therefore shape-dependent and more at the larger
models.

\begin{figure}[t]
\centering
\begin{minipage}[b]{0.24\linewidth}
\centering
\includegraphics[width=\linewidth]{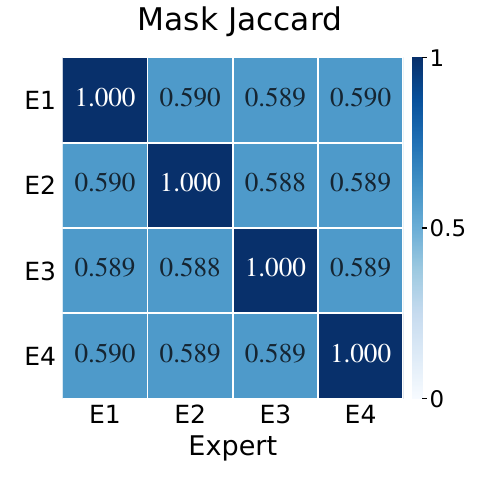}
{\scriptsize (a) Trained overlap}
\end{minipage}\hfill%
\begin{minipage}[b]{0.24\linewidth}
\centering
\includegraphics[width=\linewidth]{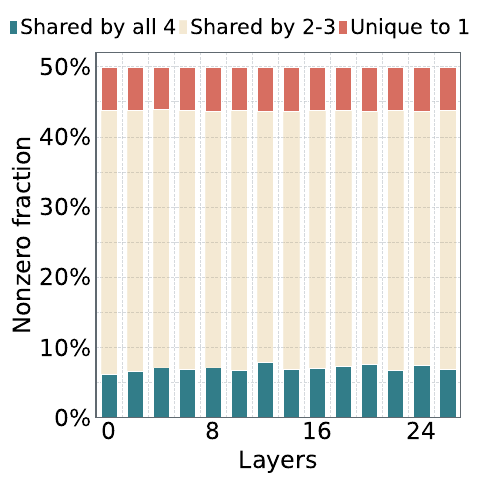}
{\scriptsize (b) Structured}
\end{minipage}\hfill%
\begin{minipage}[b]{0.24\linewidth}
\centering
\includegraphics[width=\linewidth]{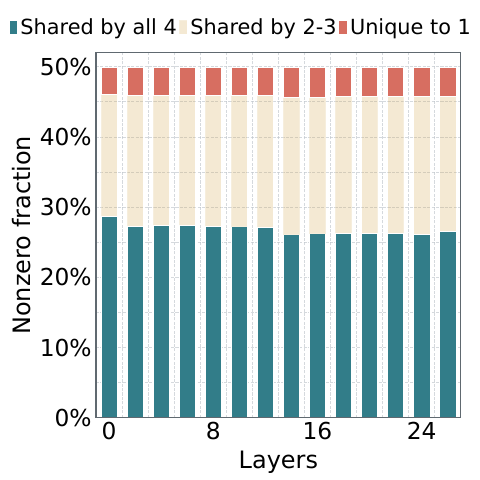}
{\scriptsize (c) Semi-structured 2:4}
\end{minipage}\hfill%
\begin{minipage}[b]{0.24\linewidth}
\centering
\includegraphics[width=\linewidth]{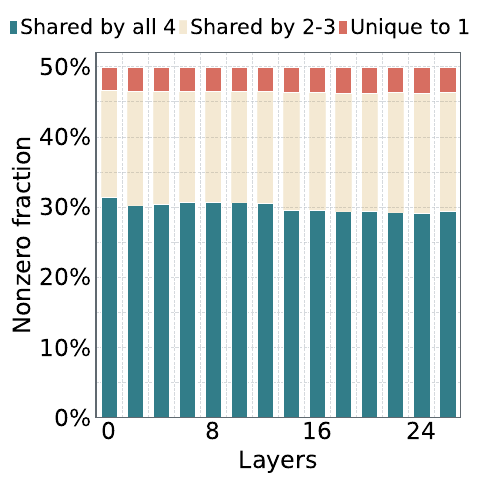}
{\scriptsize (d) Unstructured}
\end{minipage}
\caption{Learned mask structure on Qwen3-1.7B: (a) 2:4 mask Jaccard overlap; (b)-(d) layer-wise sharing of retained connections.}
\label{fig:mask-structure}
\end{figure}

\begin{figure}[t]
\centering
\includegraphics[width=\linewidth]{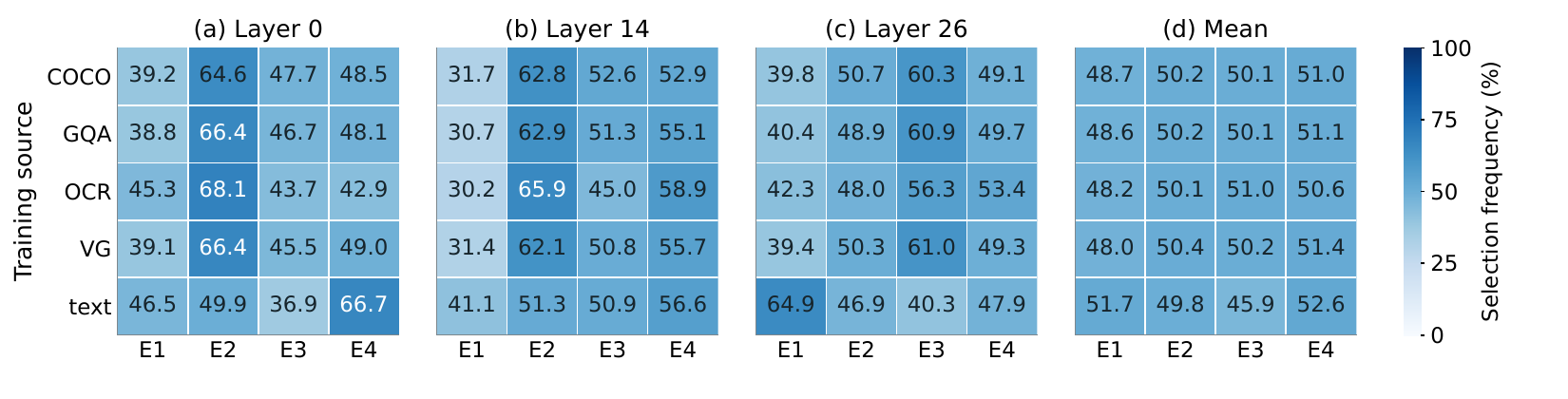}
\caption{Layer-local expert selection by training source on Qwen3-1.7B. Panels show layers 0, 14, 26, and an equal index-wise mean over all 14 converted layers. Expert IDs are not aligned across layers. 50\% indicates balanced top-2 usage.}
\label{fig:mask-domain-selection}
\end{figure}

\noindent{\textbf{Structure of learned masks.}}
Fig.~\ref{fig:mask-structure} (a) reports pairwise Jaccard overlap
between the learned 2:4 masks, defined as the intersection divided by
the union of their retained weight positions and averaged over
converted FFN projections and runs. The off-diagonal values range from
0.588 to 0.590, showing that the experts retain overlapping, but nonidentical, connectivity.

\noindent{\textbf{Latent shared expert.}} Fig.~\ref{fig:mask-structure}
(b) to (d) decompose the average retained weight density at each converted
layer into connections shared by all four experts, shared by two or
three, and unique to one. The fractions are normalized by the number of
dense FFN weight positions and averaged over experts, so each bar sums
to 50\%. The all-four intersection constitutes the \emph{latent shared
expert}: a connectivity core implicitly present in every mask. All three
mask families contain both shared and expert-specific connections, but
their compositions differ. For semi-structured and unstructured masks,
connections shared by all four experts account for more than half of the
retained entries. Neuron-structured masks have a much smaller all-four
intersection and a larger partially shared component. These profiles
remain broadly stable across the converted layers. 

\noindent{\textbf{Routing across training sources.}}
Fig.~\ref{fig:mask-domain-selection} reports expert selection
frequencies for prompts from five training-source buckets, including COCO~\citep{lin2015microsoftcococommonobjects}, GQA~\citep{hudson2018gqa},
OCR~\citep{8978122}, VG~\citep{krishna2016visualgenomeconnectinglanguage}, and text. Each entry averages binary dispatch indicators over
prefill tokens within a prompt, then equally over prompts from the
same source. We use 64 prompts per source. The three displayed layers,
0, 14, and 26, are selected by depth. The mean panel equally averages
all 14 converted layers, not only these three.
Expert IDs are local to each layer and are not aligned before averaging,
so this panel summarizes index-wise utilization, not four global experts.
With two of four experts selected per token, balanced usage
corresponds to a 50\% selection frequency per expert. Individual layers exhibit source-related usage differences, while the all-layer mean is closer to balanced usage, obscuring the larger contrasts visible at
individual layers. These observations indicate source-related, layer-local routing patterns. More detailed analysis can be found in Tab.~\ref{tab:add-functional}.

\section{Conclusion}
\label{sec:conclusion}

We presented MASKerade, a dense-to-MoE upcycling method that learns
token-routed mask experts over frozen pretrained FFN weights.
Vision-language experiments and controlled ablations support learning
expert connectivity as an effective alternative to updating independent
expert weights. The learned masks retain a shared connectivity core
alongside expert-specific connections, with routing adapting their use
to individual tokens. Future work will investigate inference kernels
that reuse overlapping computations while preserving expert-specific
nonlinear transformations, and assess their end-to-end latency and
memory trade-offs.

\newpage







\bibliography{iclr2027_conference}
\bibliographystyle{iclr2027_conference}

\newpage

\appendix
\section*{Appendix}

\section{Implementation Details}
\label{sec:protocol-details}

\noindent{\textbf{Default configuration.}}
Tab.~\ref{tab:maskerade-implementation} summarizes the MASKerade
training setup. Across backbones, alternating FFNs are converted into
four 2:4 mask experts with deterministic top-2 routing in both training
and inference. Mask scores and routers are optimized while pretrained
language-model weights remain frozen. We use importance initialization
and one epoch of AdamW training with a peak learning rate of
$5\times10^{-4}$. The small and large model settings differ in global
batch size, hardware allocation, and text length limit, as listed in the
table.

\begin{table}[!htbp]
\centering
\caption{Default MASKerade implementation details.}
\label{tab:maskerade-implementation}
\small
\setlength{\tabcolsep}{4pt}
\renewcommand{\arraystretch}{1.15}
\begin{tabular}{@{}p{0.29\linewidth}p{0.67\linewidth}@{}}
\toprule
Setting & Value \\
\midrule
Training data & 665K multimodal instruction mixture \\
Converted FFNs & Even-indexed layers: Qwen3-1.7B, 14/28; Gemma3-1B, 13/26; Qwen3.6-27B, 32/64; Gemma4-31B, 30/60 \\
Experts and routing & Four experts with deterministic top-2 in training and inference using dropless dispatch \\
Default masks & Semi-structured 2:4 \\
Trainable LM parameters & Mask scores and routers \\
Pretrained LM weights & Frozen \\
Score initialization & Importance initialization with noise scale $\alpha_{\mathrm{init}}=0.1$ \\
Gradient estimator & Straight-through estimator for hard mask selection \\
Optimizer & AdamW \\
Learning rate & $5\times10^{-4}$ for mask scores and routers \\
Training epochs & 1 \\
Learning-rate schedule & Cosine decay with 3\% warmup \\
Weight decay & 0 \\
Load-balancing coefficient & $\lambda_{\mathrm{bal}}=0.01$ \\
Precision & Bfloat16 \\
Global batch size & Small: 128 (8 NVIDIA H100 GPUs); large: 256 (64 NVIDIA B200 GPUs) \\
Text length limit & Small: 2048; large: 1,536 tokens before visual-token expansion \\
Evaluation metrics &
\begin{tabular}[t]{@{}l@{}}
GQA: accuracy (\%) \\
MME: perception score \\
POPE: mean F1 across three splits (\%) \\
SQA: image-subset accuracy (\%) \\
TextVQA: consensus VQA accuracy (\%)
\end{tabular} \\
\bottomrule
\end{tabular}
\end{table}

\section{More analysis}

\subsection{Score initialization}
\begin{figure}[t]
\centering
\includegraphics[width=\linewidth]{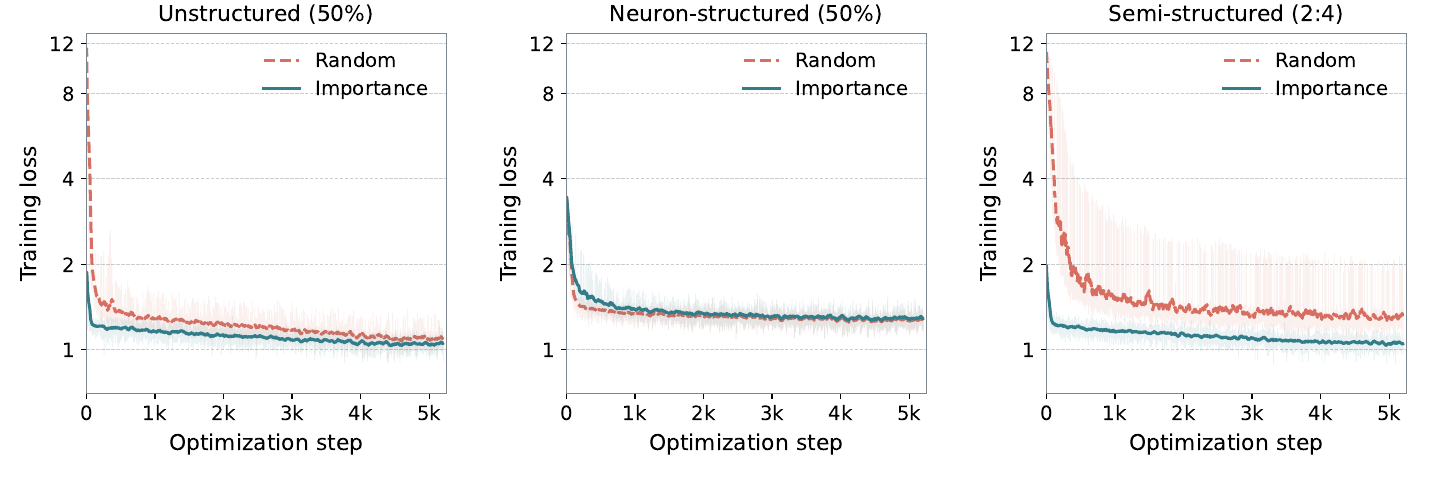}
\caption{Random versus importance initialization across mask granularities on Qwen3-1.7B.}
\label{fig:mask-initialization-all}
\end{figure}

\begin{figure}[t]
\centering
\includegraphics[width=\linewidth]{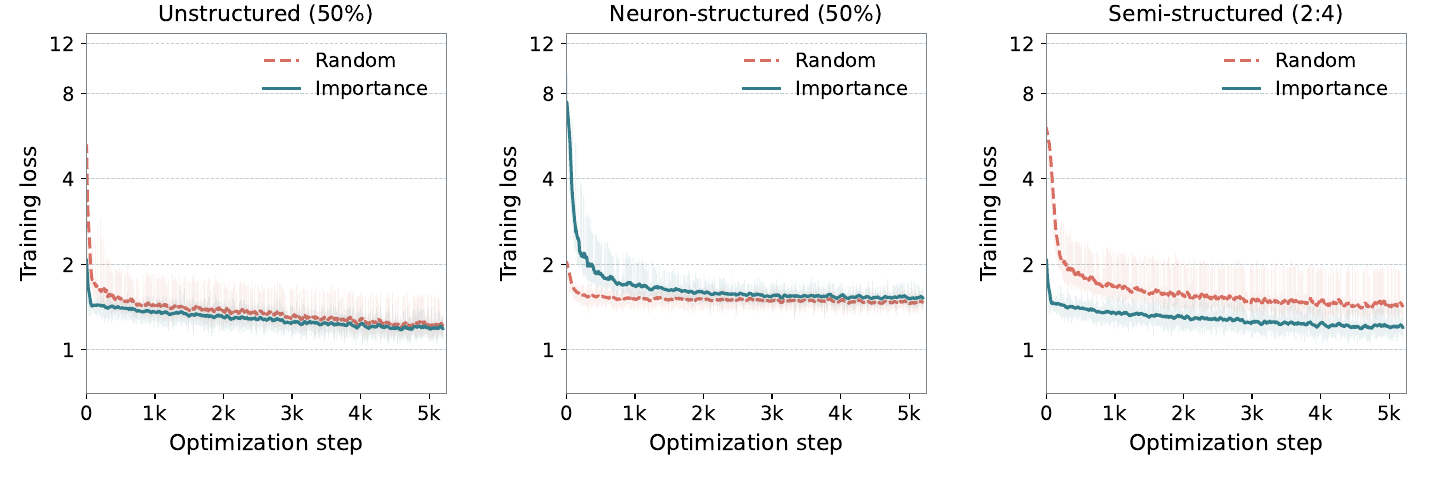}
\caption{Random versus importance initialization across mask granularities on Gemma3-1B.}
\label{fig:mask-initialization-all-gemma}
\end{figure}

\noindent{\textbf{Initialization dynamics.}}
Fig.~\ref{fig:mask-initialization-all} and
Fig.~\ref{fig:mask-initialization-all-gemma} compare training-loss
trajectories under random and importance initialization on Qwen3-1.7B
and Gemma3-1B. Importance initialization lowers the initial loss for
unstructured and 2:4 masks on both backbones. The unstructured gap
narrows during training, whereas importance-initialized 2:4 masks
retain lower loss and exhibit fewer large fluctuations. The
neuron-structured curves are closer on Qwen3-1.7B. For Gemma3-1B,
random initialization starts with substantially lower loss and remains
lower near the end of training. These curves show that the optimization
effect of initialization depends on mask granularity and backbone.

\noindent{\textbf{Score initialization on Gemma.}}
Tab.~\ref{tab:gemma-add-init} compares random, importance, and
importance-permuted initialization across the three mask granularities
on Gemma3-1B. Importance initialization has the highest reported score
on all five benchmarks within each family. Its advantage over random
initialization is particularly pronounced for 2:4 masks on GQA and
TextVQA, while the GQA difference is smaller for unstructured masks.
Permuting importance scores also reduces performance in every family,
supporting the value of aligning the initialization prior with its
original parameter locations.

\noindent{\textbf{Mask ratio on Gemma.}}
Tab.~\ref{tab:gemma-ratio} varies the per-expert retained fraction from 25\%
to 75\% across the three mask families on Gemma3-1B, with four experts
and top-2 routing. Unstructured masks lead on four benchmarks at each
retention level. Semi-structured masks achieve the highest TextVQA
scores at 25\% and 50\% retention, while neuron-structured masks
lead on TextVQA at 75\%.

\noindent{\textbf{Matched activation budgets on Gemma.}}
Tab.~\ref{tab:gemma-compmatch} compares low-budget MASKerade variants with a
single learned mask and existing MoE methods on Gemma3-1B. At 0.84B
nominal activated parameters, 1:4/top-2 achieves higher reported means
than the single 2:4 mask and all four external baselines on all five
benchmarks. The 2:4/top-1 variant exceeds the single mask on four
benchmarks, with a slightly lower GQA mean (58.24 versus 58.30).
These comparisons support learning multiple token-selectable masks
under a fixed nominal activation budget. Unlike on Qwen3-1.7B,
1:4/top-2 outperforms 2:4/top-1 on four benchmarks, with POPE as the
exception, indicating that the preferred allocation between expert
density and selection count depends on the backbone. 

\begin{figure}[t]
\centering
\begin{minipage}[b]{0.24\linewidth}
\centering
\includegraphics[width=\linewidth]{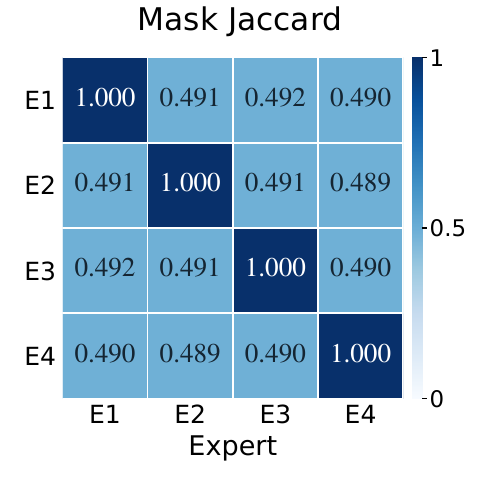}
{\scriptsize (a) Trained overlap}
\end{minipage}\hfill%
\begin{minipage}[b]{0.24\linewidth}
\centering
\includegraphics[width=\linewidth]{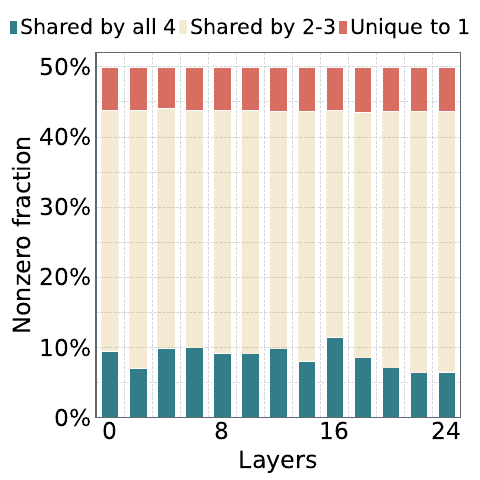}
{\scriptsize (b) Structured}
\end{minipage}\hfill%
\begin{minipage}[b]{0.24\linewidth}
\centering
\includegraphics[width=\linewidth]{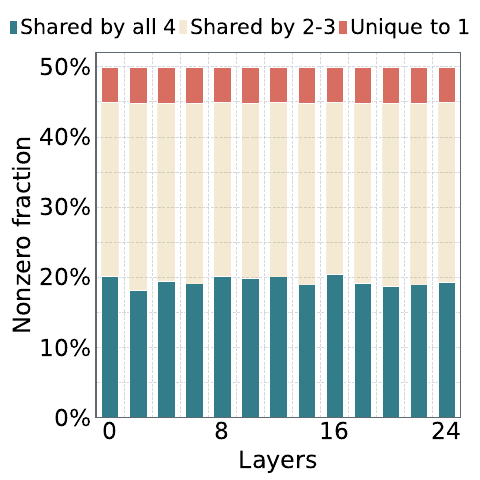}
{\scriptsize (c) Semi-structured 2:4}
\end{minipage}\hfill%
\begin{minipage}[b]{0.24\linewidth}
\centering
\includegraphics[width=\linewidth]{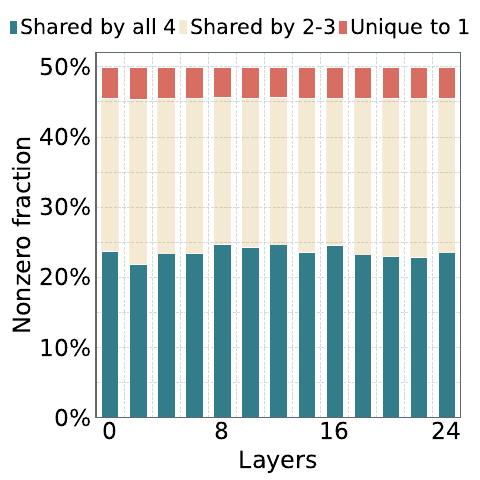}
{\scriptsize (d) Unstructured}
\end{minipage}
\caption{Learned mask structure on Gemma3-1B, averaged over three runs per family: (a) 2:4 Jaccard overlap; (b)-(d) layer-wise sharing of retained connections, normalized by dense weight positions.}
\label{fig:gemma-mask-structure}
\end{figure}

\noindent{\textbf{Structure of learned masks on Gemma.}}
Fig.~\ref{fig:gemma-mask-structure} (a) reports pairwise Jaccard overlap
between learned 2:4 masks in the Gemma3-1B runs, averaged over
converted FFN projections and runs. The off-diagonal values range
from 0.489 to 0.492, indicating overlapping but nonidentical retained
connections. The similar overlap across expert pairs suggests that
no pair is substantially more alike than the others under this
structural measure. This establishes connectivity diversity, while
the usefulness of the routed assignments requires separate
functional and intervention analyses.

\noindent{\textbf{Latent shared expert on Gemma.}}
Fig.~\ref{fig:gemma-mask-structure}
(b) to (d) decompose the average retained weight density at each converted
layer into connections shared by all four experts, shared by two or
three, and unique to one. Each bar sums to 50\% of dense weight
positions. The all-four intersection forms a \emph{latent shared
expert}, with a smaller shared core for neuron-structured masks than
for either weight-level family. In these Gemma3-1B runs,
the all-four component accounts for less than half of retained entries
in every family. All three constructions therefore combine a common
connectivity core with partially shared and expert-specific connections.

\begin{figure}[!htbp]
\centering
\includegraphics[width=\linewidth]{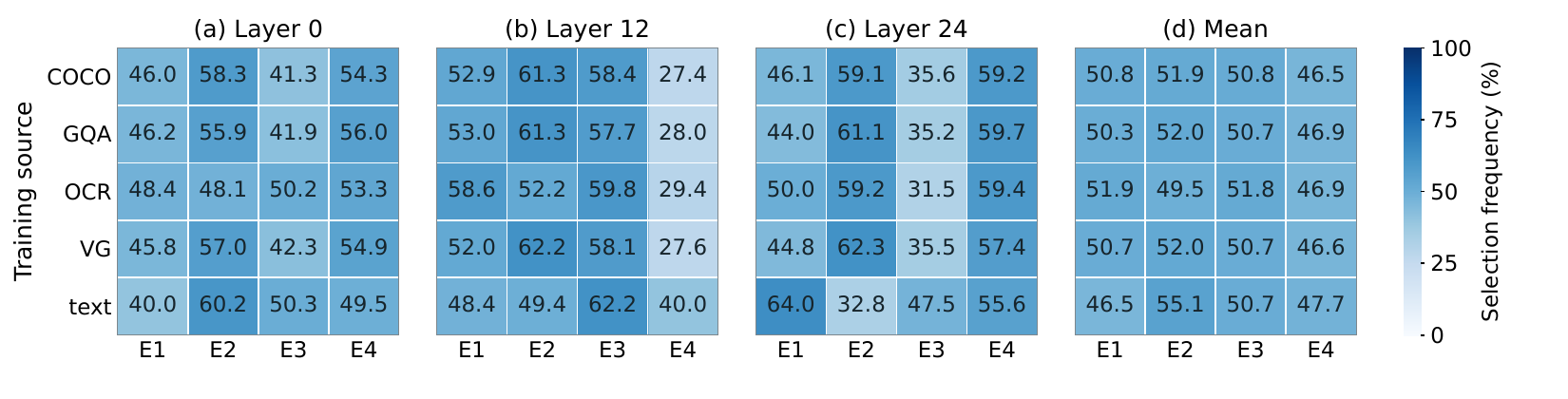}
\caption{Layer-local expert selection by training source on Gemma3-1B. Panels show layers 0, 12, 24, and the unaligned index-wise mean over all 13 converted layers. 50\% denotes balanced top-2 usage.}
\label{fig:gemma-mask-domain-selection}
\end{figure}

\noindent{\textbf{Routing across training sources.}}
Fig.~\ref{fig:gemma-mask-domain-selection} reports expert selection
frequencies for five training-source buckets on Gemma3-1B. Each entry averages token selection frequencies within
a prompt and then equally over 64 prompts per source. Layers 0, 12,
and 24 show source-related differences, with text prompts exhibiting
a distinct selection pattern at layer 24. The mean over all 13
converted layers is closer to the balanced 50\% usage level.

\begin{table}[t]
\centering
\renewcommand{\arraystretch}{0.95}
\begin{minipage}[t]{0.48\linewidth}
\centering
\captionof{table}{Mask-score initialization on Gemma3-1B (four experts, top-2, 50\% retention).}
\label{tab:gemma-add-init}
\scriptsize
\setlength{\tabcolsep}{2pt}
\begin{adjustbox}{max width=\linewidth}
\begin{tabular}{lccccc}
\toprule
Initialization & GQA $\uparrow$ & MME $\uparrow$ & POPE $\uparrow$ & SQA $\uparrow$ & TextVQA $\uparrow$ \\
\midrule
\multicolumn{6}{l}{\emph{Structured 50\%}} \\
Random & 53.13 & 1004.9 & 81.00 & 50.10 & 39.58 \\
Importance & 56.65 & 1251.4 & 85.68 & 51.74 & 41.80 \\
Importance-permuted & 53.16 & 1008.2 & 82.76 & 49.85 & 40.20 \\
\midrule
\multicolumn{6}{l}{\emph{Unstructured 50\%}} \\
Random & 59.92 & 1210.7 & 85.72 & 46.65 & 38.85 \\
Importance & 60.46 & 1295.3 & 86.39 & 56.64 & 42.73 \\
Importance-permuted & 59.34 & 1221.0 & 85.41 & 52.74 & 39.49 \\
\midrule
\multicolumn{6}{l}{\emph{Semi-structured 2:4}} \\
Random & 51.54 & 1068.2 & 83.37 & 45.60 & 26.89 \\
Importance & 60.34 & 1294.4 & 86.35 & 55.40 & 42.81 \\
Importance-permuted & 58.89 & 1203.8 & 85.27 & 42.94 & 37.07 \\
\bottomrule
\end{tabular}
\end{adjustbox}
\end{minipage}
\hfill
\begin{minipage}[t]{0.48\linewidth}
\centering
\captionof{table}{Per-expert retention across mask granularities on Gemma3-1B (four experts, top-2).}
\label{tab:gemma-ratio}
\scriptsize
\setlength{\tabcolsep}{2pt}
\begin{adjustbox}{max width=\linewidth}
\begin{tabular}{lccccc}
\toprule
Family & GQA $\uparrow$ & MME $\uparrow$ & POPE $\uparrow$ & SQA $\uparrow$ & TextVQA $\uparrow$ \\
\midrule
\multicolumn{6}{l}{\emph{Structured}} \\
25\% FFN parameters & 51.25 & 1105.5 & 83.01 & 48.86 & 39.03 \\
50\% FFN parameters & 56.65 & 1251.4 & 85.68 & 51.74 & 41.80 \\
75\% FFN parameters & 58.95 & 1292.8 & 85.09 & 51.47 & 45.21 \\
\midrule
\multicolumn{6}{l}{\emph{Unstructured}} \\
25\% FFN parameters & 59.82 & 1256.2 & 85.40 & 54.22 & 41.74 \\
50\% FFN parameters & 60.46 & 1295.3 & 86.39 & 56.64 & 42.73 \\
75\% FFN parameters & 60.78 & 1315.8 & 86.31 & 56.48 & 43.02 \\
\midrule
\multicolumn{6}{l}{\emph{Semi-structured}} \\
1:4 mask & 59.09 & 1233.7 & 85.11 & 52.99 & 41.87 \\
2:4 mask & 60.34 & 1294.4 & 86.35 & 55.40 & 42.81 \\
3:4 mask & 60.37 & 1304.4 & 86.12 & 55.19 & 43.21 \\
\bottomrule
\end{tabular}
\end{adjustbox}
\end{minipage}
\end{table}

\begin{table}[t]
\caption{Matched nominal activation budgets on Gemma3-1B. Calculation denotes retained density per expert times the number of activated experts. Only protocol-matched runs are included.}
\label{tab:gemma-compmatch}
\begin{center}
\small
\resizebox{\textwidth}{!}{%
\begin{tabular}{lcccccccc}
\toprule
Configuration & Masks & Calculation & Act.\ params & GQA $\uparrow$ & MME $\uparrow$ & POPE $\uparrow$ & SQA $\uparrow$ & TextVQA $\uparrow$ \\
\midrule

MoEfication~\citep{zhang-etal-2022-moefication}
& -
& -
& 0.84B
& \ms{55.36}{0.28}
& \ms{1212.9}{13.5}
& \ms{80.92}{0.35}
& \ms{50.33}{0.55}
& \ms{40.68}{0.58} \\

LLaMA-MoE~\citep{zhu-etal-2024-llama}
& -
& -
& 0.84B
& \ms{57.67}{0.33}
& \ms{1234.7}{11.2}
& \ms{85.00}{0.42}
& \ms{52.71}{0.63}
& \ms{40.96}{0.67} \\

DIVE~\citep{feng-etal-2025-dive}
& -
& -
& 0.85B
& \ms{51.66}{0.27}
& \ms{1129.3}{13.2}
& \ms{82.80}{0.28}
& \ms{50.84}{0.63}
& \ms{41.79}{0.58} \\

ToMoE~\citep{gao2026tomoe}
& -
& -
& 0.84B
& \ms{53.96}{0.15} 
& \ms{1123.0}{17.9} 
& \ms{81.37}{0.43} 
& \ms{51.98}{0.52} 
& \ms{40.63}{0.32} \\

\cmidrule{1-9}

1 mask only
& 2:4
& $50\%\times1$
& 0.84B
& \ms{58.30}{0.30}
& \ms{1239.9}{13.7}
& \ms{84.87}{0.38}
& \ms{52.59}{0.69}
& \ms{40.07}{0.68} \\

{[Ours]} 4e/top-2
& 1:4
& $25\%\times2$
& 0.84B
& \ms{59.09}{0.33}
& \ms{1248.7}{15.5}
& \ms{85.11}{0.36}
& \ms{52.99}{0.61}
& \ms{42.47}{0.64} \\

{[Ours]} 4e/top-1
& 2:4
& $50\%\times1$
& 0.84B
& \ms{58.24}{0.31}
& \ms{1246.8}{11.3}
& \ms{85.77}{0.35}
& \ms{52.61}{0.61}
& \ms{41.46}{0.67} \\

{[Ours]} 4e/top-2
& 2:4
& $50\%\times2$
& 1.00B
& \ms{60.07}{0.27}
& \ms{1293.8}{11.9}
& \ms{86.41}{0.10}
& \ms{55.30}{0.47}
& \ms{42.90}{0.47} \\

\bottomrule
\end{tabular}}
\end{center}
\end{table}

\noindent{\textbf{Updating shared FFN weights.}}
Tab.~\ref{tab:trainw} examines whether learning the underlying FFN
weights improves mask-based adaptation. On Qwen3-1.7B, weight updates
improve all five benchmarks for a single mask and four benchmarks
for 4e/top-2. For Gemma3-1B, weight updates improve all five benchmarks
in both configurations.
The frozen-weight multi-expert configuration also exceeds Wanda
prune-then-tune on all five benchmarks on Qwen3-1.7B and
Gemma3-1B. Weight adaptation can therefore
complement mask learning.

\begin{table}[t]
\caption{Mask learning with frozen or trainable shared FFN weights on Qwen3-1.7B and Gemma3-1B.}
\label{tab:trainw}
\centering
\small
\setlength{\tabcolsep}{4pt}

\begin{adjustbox}{max width=\linewidth}
\begin{tabular}{clccccc}
\toprule
& Configuration & GQA {↑} & MME {↑} & POPE {↑} & SQA {↑} & TextVQA {↑} \\
\midrule

\multirow{5}{*}{\rotatebox{90}{Qwen3-1.7B}}
& learned mask only (1 mask)
& \ms{56.51}{0.32} 
& \ms{1258.1}{15.5}
& \ms{84.50}{0.25}
& \ms{59.67}{0.56}
& \ms{49.13}{0.68} \\

& learned mask + trained weights (1 mask)
& \ms{60.69}{0.05} 
& \ms{1310.8}{0.4} 
& \ms{85.93}{0.39}
& \ms{61.41}{0.48} 
& \ms{50.49}{0.35} \\

& learned mask only (4e/top-2)
& \ms{62.00}{0.28}
& \ms{1332.4}{11.9}
& \ms{86.49}{0.16}
& \ms{63.48}{0.57}
& \ms{52.50}{0.37} \\

& learned masks + trained weights (4e/top-2)
& \ms{62.73}{0.33} 
& \ms{1353.2}{1.2} 
& \ms{86.46}{0.54}
& \ms{63.59}{0.52} 
& \ms{52.93}{0.24} \\

& prune-then-tune (Wanda, 4e/top-2)
& \ms{58.27}{0.08} 
& \ms{1282.7}{20.4} 
& \ms{85.81}{0.40} 
& \ms{61.29}{0.52} 
& \ms{49.88}{0.17} \\

\midrule

\multirow{5}{*}{\rotatebox{90}{Gemma3-1B}}
& learned mask only (1 mask)
& \ms{58.30}{0.30}
& \ms{1239.9}{13.7}
& \ms{84.87}{0.38}
& \ms{52.59}{0.69}
& \ms{40.07}{0.68} \\

& learned mask + trained weights (1 mask)
& \ms{60.06}{0.22}
& \ms{1240.9}{4.6}
& \ms{85.49}{0.17}
& \ms{52.66}{0.37}
& \ms{40.70}{0.38} \\

& learned mask only (4e/top-2)
& \ms{60.07}{0.27}
& \ms{1293.8}{11.9}
& \ms{86.41}{0.10}
& \ms{55.30}{0.47}
& \ms{42.90}{0.47} \\

& learned masks + trained weights (4e/top-2)
& \ms{60.95}{0.33}
& \ms{1296.7}{4.1}
& \ms{86.53}{0.16}
& \ms{55.90}{0.46}
& \ms{44.70}{0.43} \\

& prune-then-tune (Wanda, 4e/top-2)
& \ms{58.25}{0.23}
& \ms{1255.0}{13.7}
& \ms{84.69}{0.47}
& \ms{51.81}{0.45}
& \ms{37.01}{0.34} \\

\bottomrule
\end{tabular}
\end{adjustbox}
\end{table}

\noindent{\textbf{Number of activated experts.}}
Tab.~\ref{tab:nexperts} and Tab.~\ref{tab:gemma-nexperts} vary top-$k$
while keeping four 2:4 experts. These are configuration-level comparisons:
Eq.~\ref{eq:routed-output} retains the selected probability for top-1
but renormalizes the selected probabilities for $k\geq2$.
For Qwen3-1.7B, activating all four
experts gives the highest GQA, SQA, and TextVQA scores, whereas
top-2 performs best on MME and POPE. For Gemma3-1B, top-2 leads on
four benchmarks and top-3 gives a slightly higher MME score.
Activating all four experts lowers every Gemma score relative to
top-2 despite doubling the nominal selected-expert FFN computation.
Since each expert retains half of the FFN weights, increasing $k$
from one to four raises converted-FFN computation from half to twice
that of a dense pass without guaranteeing better performance.


\begin{table}[t]
    \centering
    \renewcommand{\arraystretch}{0.95}
    \setlength{\tabcolsep}{2pt}

    \begin{minipage}[t]{0.48\textwidth}
        \centering
        \captionof{table}{Varying top-$k$ with four 2:4 experts on Qwen3-1.7B.}
        \label{tab:nexperts}

        \scriptsize
        \scalebox{1.2}{
        \begin{tabular}{cccccc}
            \toprule
            Top-$k$ & GQA {↑} & MME {↑} & POPE {↑} & SQA {↑} & TextVQA {↑} \\
            \midrule
            1 & 62.20 & 1323.1 & 86.31 & 63.10 & 51.59 \\
            2 & 62.08 & \textbf{1335.1} & \textbf{86.63} & 63.99 & 52.86 \\
            3 & 62.59 & 1297.8 & 86.32 & 64.30 & 52.52 \\
            4 & \textbf{62.75} & 1309.0 & 86.32 & \textbf{64.40} & \textbf{53.07} \\
            \bottomrule
        \end{tabular}
        }
    \end{minipage}
    \hfill
    \begin{minipage}[t]{0.48\textwidth}
        \centering
        \captionof{table}{Varying the number of 2:4 experts with top-1 routing on Qwen3-1.7B.}
        \label{tab:expertcount}

        \scriptsize
        \scalebox{1.2}{
        \begin{tabular}{cccccc}
            \toprule
            $N$ & GQA {↑} & MME {↑} & POPE {↑} & SQA {↑} & TextVQA {↑} \\
            \midrule
            1 & 56.36 & 1265.6 & 84.60 & 60.07 & 49.73 \\
            2 & 60.46 & 1304.4 & 85.98 & 61.87 & 50.99 \\
            4 & \textbf{62.20} & 1323.1 & 86.31 & \textbf{63.10} & \textbf{51.59} \\
            8 & 61.87 & \textbf{1330.5} & \textbf{86.50} & 63.04 & 51.55 \\
            \bottomrule
        \end{tabular}
        }
    \end{minipage}
\end{table}

\begin{table}[!htbp]
\centering
\renewcommand{\arraystretch}{0.95}
\setlength{\tabcolsep}{2pt}
\begin{minipage}[t]{0.48\textwidth}
\centering
\captionof{table}{Varying top-$k$ with four 2:4 experts on Gemma3-1B.}
\label{tab:gemma-nexperts}
\scriptsize
\scalebox{1.2}{
\begin{tabular}{cccccc}
\toprule
Top-$k$ & GQA $\uparrow$ & MME $\uparrow$ & POPE $\uparrow$ & SQA $\uparrow$ & TextVQA $\uparrow$ \\
\midrule
1 & 58.24 & 1246.8 & 85.77 & 52.11 & 40.86 \\
2 & \textbf{60.34} & 1294.4 & \textbf{86.35} & \textbf{55.40} & \textbf{42.81} \\
3 & 59.72 & \textbf{1295.9} & 84.93 & 53.00 & 39.39 \\
4 & 60.10 & 1267.6 & 85.25 & 54.49 & 41.48 \\
\bottomrule
\end{tabular}
}
\end{minipage}
\hfill
\begin{minipage}[t]{0.48\textwidth}
\centering
\captionof{table}{Varying the number of 2:4 experts with top-1 routing on Gemma3-1B.}
\label{tab:gemma-expertcount}
\scriptsize
\scalebox{1.2}{
\begin{tabular}{cccccc}
\toprule
$N$ & GQA $\uparrow$ & MME $\uparrow$ & POPE $\uparrow$ & SQA $\uparrow$ & TextVQA $\uparrow$ \\
\midrule
1 & 58.50 & 1237.4 & 84.87 & \textbf{52.64} & 40.57 \\
2 & \textbf{59.10} & \textbf{1301.7} & 85.79 & 50.72 & 40.27 \\
4 & 58.24 & 1246.8 & 85.77 & 52.11 & \textbf{40.86} \\
8 & 57.70 & 1299.4 & \textbf{85.91} & 51.96 & 36.47 \\
\bottomrule
\end{tabular}
}
\end{minipage}
\end{table}

\noindent{\textbf{Number of available experts.}}
Tab.~\ref{tab:expertcount} and Tab.~\ref{tab:gemma-expertcount} vary
the size of the mask bank while selecting one 2:4 expert per token.
For Qwen3-1.7B, moving from one to four experts improves all five
benchmarks. Eight experts further improve MME and POPE but reduce
the other three scores. For Gemma3-1B, two experts give the highest
GQA and MME scores, while one expert leads on SQA, four on TextVQA,
and eight on POPE. Larger mask banks therefore do not yield uniform
gains across backbones or tasks. Selected-expert FFN computation
remains fixed in this comparison, although the number of mask
scores and the router size increase with the expert bank.
The single-mask case bypasses routing, whereas multi-expert top-1
uses probability-weighted outputs. The comparison therefore does not
isolate expert-bank size from output scaling.

\noindent{\textbf{Placement of converted layers.}}
Tab.~\ref{tab:placement} compares where we introduce mask experts.
On Qwen3-1.7B, even-indexed layers give the highest MME and TextVQA
scores, while the first and second halves lead on POPE and SQA,
respectively. For Gemma3-1B, even-indexed layers lead on MME, POPE,
and TextVQA, and odd-indexed layers give the highest SQA score.
Converting every FFN gives the highest GQA score on both backbones,
but lowers the other four scores relative to even-layer conversion.
These results support alternating even layers as a practical default
without requiring conversion of the entire FFN stack.

\begin{table}[t]
\caption{Effect of MoE layer placement across model backbones (four 2:4 experts, top-2).}
\label{tab:placement}
\centering
\small
\setlength{\tabcolsep}{4pt}

\begin{adjustbox}{max width=\linewidth}
\begin{tabular}{clcccccc}
\toprule
& Placement & MoE layers & GQA {↑} & MME {↑} & POPE {↑} & SQA {↑} & TextVQA {↑} \\
\midrule

\multirow{5}{*}{\rotatebox{90}{Qwen3-1.7B}}
& Even layers
& 14
& 62.08
& \textbf{1335.1}
& 86.63
& 63.99
& \textbf{52.86} \\

& Odd layers
& 14
& 62.14
& 1326.6
& 85.20
& 61.92
& 51.69 \\

& First half
& 14
& 61.26
& 1296.0
& \textbf{86.73}
& 58.30
& 50.92 \\

& Second half
& 14
& 60.76
& 1299.2
& 85.96
& \textbf{67.23}
& 50.42 \\

& All layers
& 28
& \textbf{62.47}
& 1298.3
& 86.58
& 60.44
& 50.02 \\

\midrule

\multirow{5}{*}{\rotatebox{90}{Gemma3-1B}}
& Even layers
& 13
& 60.34 
& \textbf{1294.4} 
& \textbf{86.35} 
& 55.40 
& \textbf{42.81} \\

& Odd layers
& 13
& 60.38
& 1245.5
& 85.85
& \textbf{56.12}
& 40.76 \\

& First half
& 13
& 59.33
& 1236.3
& 85.05
& 53.15
& 34.11 \\

& Second half
& 13
& 58.74
& 1154.9
& 84.91
& 50.62
& 34.29 \\

& All layers
& 26
& \textbf{60.57}
& 1192.4
& 85.78
& 46.80
& 37.90 \\

\bottomrule
\end{tabular}
\end{adjustbox}
\end{table}

\noindent{\textbf{Functional diversity of mask experts.}}
Tab.~\ref{tab:add-functional} uses 600 examples per backbone:
300 GQA test-dev and 300 TextVQA validation examples.
Identical token hidden states are fed to all four experts. Pairwise
statistics average tokens within each prompt, then equally across
prompts and the six expert pairs, separately for vision, prompt, and
teacher-forced response tokens (including the template separator).
Mean pairwise cosine similarities range from 0.65 to 0.90,
with positive normalized distances throughout, showing that the masks
induce different transformations rather than identical expert outputs.
For both backbones, the middle probed layer exhibits lower cosine
similarity and greater normalized distance than the early and late
layers across all three token types. This complements the connectivity
overlap analysis in Fig.~\ref{fig:mask-structure}: shared weight
positions can coexist with functionally distinct outputs. 

\begin{table}[!htbp]
\centering\small
\setlength{\tabcolsep}{4pt}
\caption{Identical-input expert outputs by token type. Cosine and Distance are pairwise averages. Distance normalizes $L_2$ differences by summed output norms, and $e_i$ denotes the mean output norm.}
\label{tab:add-functional}
\begin{tabular}{llrrrrrr}
\toprule
Layer & Token type & Cosine & Distance & $e_0$ & $e_1$ & $e_2$ & $e_3$ \\
\midrule

\multicolumn{8}{l}{\emph{Qwen3-1.7B}} \\
0  & Vision   & 0.700 & 0.390 & 13.60 & 13.70 & 13.50 & 13.80 \\
0  & Prompt   & 0.780 & 0.330 &  9.20 &  9.30 &  9.10 &  9.40 \\
0  & Response & 0.720 & 0.370 & 11.60 & 11.40 & 11.80 & 12.00 \\

14 & Vision   & 0.650 & 0.420 & 52.00 & 53.00 & 51.00 & 54.00 \\
14 & Prompt   & 0.730 & 0.370 & 44.00 & 45.00 & 46.00 & 44.50 \\
14 & Response & 0.670 & 0.410 & 47.00 & 49.00 & 48.00 & 48.50 \\

26 & Vision   & 0.780 & 0.330 & 355.0 & 360.0 & 350.0 & 365.0 \\
26 & Prompt   & 0.800 & 0.320 & 510.0 & 520.0 & 500.0 & 515.0 \\
26 & Response & 0.760 & 0.350 & 515.0 & 525.0 & 505.0 & 520.0 \\

\midrule

\multicolumn{8}{l}{\emph{Gemma3-1B}} \\
0  & Vision   & 0.880 & 0.240 & 24.50 & 24.30 & 26.70 & 27.00 \\
0  & Prompt   & 0.820 & 0.300 & 23.70 & 23.40 & 23.60 & 24.20 \\
0  & Response & 0.760 & 0.380 & 19.50 & 16.80 & 19.20 & 26.80 \\

12 & Vision   & 0.680 & 0.390 & 2.08 & 2.00 & 2.20 & 1.91 \\
12 & Prompt   & 0.720 & 0.360 & 2.10 & 2.12 & 2.08 & 1.92 \\
12 & Response & 0.660 & 0.410 & 2.03 & 2.08 & 2.14 & 1.95 \\

24 & Vision   & 0.900 & 0.220 & 2.64 & 2.67 & 2.62 & 2.60 \\
24 & Prompt   & 0.880 & 0.240 & 2.58 & 2.64 & 2.57 & 2.54 \\
24 & Response & 0.890 & 0.230 & 2.94 & 2.91 & 2.88 & 2.90 \\

\bottomrule
\end{tabular}
\end{table}

\noindent{\textbf{Local expert replacement.}}
Tab.~\ref{tab:add-local-replacement} uses the same 600-example subset.
At a single layer, one selected expert is replaced at every answer-predictive
position, starting with the last prompt token, while retaining its routing
weight. NLL uses teacher-forced reference answers, excluding the final
labeled token. Diagnostic accuracy uses normalized exact match after
greedy decoding (up to eight new tokens). TextVQA accepts any annotated
answer rather than using the official consensus score.
Every reported intervention increases mean and median NLL and reduces
diagnostic accuracy.
Loss increases affect 57--78\% of evaluated inputs, and correct-to-wrong
answer flips consistently exceed wrong-to-correct flips. The effects
are largest at the middle probed layers and when replacing the
first-ranked selected expert, although that slot also retains the
larger routing weight. Together with Tab.~\ref{tab:swap}, these results
support the usefulness of the learned assignments on this subset:
selected masks are not freely interchangeable with unselected ones.

\begin{table}[!htbp]
\centering\small
\setlength{\tabcolsep}{3pt}
\caption{Single-layer expert replacement with routing weights fixed. Slot and Alt.\ give zero-based router ranks among selected and unselected experts. $\Delta$ compares with original routing. C/W denote correct/incorrect answers. Accuracy changes and flips are in percentage points and percent, respectively.}
\label{tab:add-local-replacement}
\begin{tabular}{ccccccrrr}
\toprule
Layer & Slot & Alt. & Mean $\Delta$NLL & Median & $\Delta$NLL$>0$ (\%) & $\Delta$Acc & C$\to$W & W$\to$C \\
\midrule

\multicolumn{9}{l}{\emph{Qwen3-1.7B}} \\
0  & 0 & 0 & +0.025 & +0.008 & 62.0 & -1.00 & 1.50 & 0.50 \\
0  & 0 & 1 & +0.040 & +0.012 & 67.0 & -1.50 & 2.00 & 0.50 \\
0  & 1 & 0 & +0.015 & +0.004 & 60.0 & -0.50 & 1.00 & 0.50 \\
0  & 1 & 1 & +0.020 & +0.006 & 63.0 & -1.00 & 1.50 & 0.50 \\

14 & 0 & 0 & +0.080 & +0.035 & 73.0 & -3.00 & 4.00 & 1.00 \\
14 & 0 & 1 & +0.110 & +0.045 & 78.0 & -4.00 & 5.00 & 1.00 \\
14 & 1 & 0 & +0.040 & +0.015 & 66.0 & -1.50 & 2.50 & 1.00 \\
14 & 1 & 1 & +0.055 & +0.022 & 70.0 & -2.00 & 3.00 & 1.00 \\

26 & 0 & 0 & +0.060 & +0.027 & 72.0 & -2.50 & 3.50 & 1.00 \\
26 & 0 & 1 & +0.090 & +0.038 & 76.0 & -3.50 & 4.50 & 1.00 \\
26 & 1 & 0 & +0.025 & +0.009 & 64.0 & -1.00 & 2.00 & 1.00 \\
26 & 1 & 1 & +0.035 & +0.013 & 67.0 & -1.50 & 2.50 & 1.00 \\

\midrule

\multicolumn{9}{l}{\emph{Gemma3-1B}} \\
0  & 0 & 0 & +0.018 & +0.005 & 59.0 & -0.67 & 1.17 & 0.50 \\
0  & 0 & 1 & +0.030 & +0.009 & 64.0 & -1.00 & 1.50 & 0.50 \\
0  & 1 & 0 & +0.012 & +0.003 & 57.0 & -0.50 & 1.00 & 0.50 \\
0  & 1 & 1 & +0.017 & +0.005 & 60.0 & -0.67 & 1.17 & 0.50 \\

12 & 0 & 0 & +0.060 & +0.022 & 69.0 & -2.17 & 3.00 & 0.83 \\
12 & 0 & 1 & +0.075 & +0.030 & 74.0 & -2.67 & 3.50 & 0.83 \\
12 & 1 & 0 & +0.032 & +0.010 & 63.0 & -1.00 & 1.67 & 0.67 \\
12 & 1 & 1 & +0.045 & +0.016 & 67.0 & -1.50 & 2.17 & 0.67 \\

24 & 0 & 0 & +0.045 & +0.016 & 67.0 & -1.50 & 2.50 & 1.00 \\
24 & 0 & 1 & +0.060 & +0.024 & 71.0 & -2.00 & 3.00 & 1.00 \\
24 & 1 & 0 & +0.022 & +0.006 & 61.0 & -0.67 & 1.50 & 0.83 \\
24 & 1 & 1 & +0.030 & +0.010 & 64.0 & -1.00 & 1.83 & 0.83 \\

\bottomrule
\end{tabular}
\end{table}




\noindent{\textbf{Hyperparameter sensitivity.}}
Tab.~\ref{tab:add-hyperparameters} varies initialization noise and
learning rate around the default configuration. On Qwen3-1.7B,
removing initialization noise lowers all five scores. Increasing
the noise to 0.3 improves GQA, MME, and POPE but lowers SQA and
TextVQA. For Gemma3-1B, the default has the highest GQA, MME, and
POPE scores, while removing noise slightly improves SQA and using
a learning rate of $1\times10^{-4}$ slightly improves TextVQA.
The higher learning rate of $1\times10^{-3}$ lowers all five Gemma
scores. No tested alternative improves every benchmark over the
default, indicating task-dependent trade-offs around the chosen recipe.

\begin{table}[!htbp]
\centering
\small
\setlength{\tabcolsep}{3pt}
\caption{Sensitivity to initialization noise $\alpha_{\rm init}$ and learning rate across model backbones (four 2:4 experts, top-2).}
\label{tab:add-hyperparameters}

\begin{adjustbox}{max width=\linewidth}
\begin{tabular}{cccccccc}
\toprule
& $\alpha_{\rm init}$ & Learning rate & GQA {↑} & MME {↑} & POPE {↑} & SQA {↑} & TextVQA {↑} \\
\midrule

\multirow{5}{*}{\rotatebox{90}{Qwen3-1.7B}}
& 0.1 & $5\times10^{-4}$ & 62.08 & 1335.1 & 86.63 & 63.99 & 52.86 \\
& 0   & $5\times10^{-4}$ & 61.96 & 1324.5 & 86.34 & 61.48 & 52.11 \\
& 0.3 & $5\times10^{-4}$ & 62.20 & 1343.1 & 86.73 & 63.81 & 52.26 \\
& 0.1 & $1\times10^{-4}$ & 61.49 & 1287.2 & 85.81 & 65.15 & 51.92 \\
& 0.1 & $1\times10^{-3}$ & 61.89 & 1321.6 & 86.78 & 60.98 & 52.45 \\

\midrule

\multirow{5}{*}{\rotatebox{90}{Gemma3-1B}}
& 0.1 & $5\times10^{-4}$ & 60.34 & 1294.4 & 86.35 & 55.40 & 42.81 \\
& 0   & $5\times10^{-4}$ & 59.74 & 1291.5 & 85.51 & 55.78 & 42.12 \\
& 0.3 & $5\times10^{-4}$ & 59.63 & 1288.2 & 84.87 & 52.75 & 41.40 \\
& 0.1 & $1\times10^{-4}$ & 59.90 & 1251.4 & 85.87 & 53.30 & 42.99 \\
& 0.1 & $1\times10^{-3}$ & 59.13 & 1222.1 & 84.78 & 52.31 & 35.38 \\

\bottomrule
\end{tabular}
\end{adjustbox}
\end{table}

\noindent{\textbf{Training-time routing policy.}}
Tab.~\ref{tab:gate} compares deterministic top-2 training with a control
that selects the first expert by argmax and samples the second using
Gumbel-perturbed logits. Both models use deterministic routing at
evaluation, so the comparison concerns training-time expert selection.
Deterministic top-2 has higher reported means on all five benchmarks
for both backbones. The gains are larger on Gemma3-1B, particularly
on MME and SQA, whereas the Qwen3-1.7B differences on GQA and TextVQA
are small. This comparison supports the deterministic default,
without implying statistically significant improvements on every
benchmark or a need for stochastic expert selection during mask learning.

\begin{table}[t]
\caption{Training-gate comparison across model backbones (four 2:4 experts). Both settings use deterministic evaluation.}
\label{tab:gate}
\centering
\small
\setlength{\tabcolsep}{4pt}

\begin{adjustbox}{max width=\linewidth}
\begin{tabular}{lccccc}
\toprule
Training router & GQA {↑} & MME {↑} & POPE {↑} & SQA {↑} & TextVQA {↑} \\
\midrule

\multicolumn{6}{l}{\emph{Qwen3-1.7B}} \\

Argmax $+$ Gumbel
& \ms{61.97}{0.27}
& \ms{1309.7}{10.6}
& \ms{86.35}{0.10}
& \ms{63.20}{0.45}
& \ms{52.46}{0.23} \\

Deterministic top-2
& \ms{62.00}{0.28}
& \ms{1332.4}{11.9}
& \ms{86.49}{0.16}
& \ms{63.48}{0.57}
& \ms{52.50}{0.37} \\

\midrule

\multicolumn{6}{l}{\emph{Gemma3-1B}} \\

Argmax $+$ Gumbel
& \ms{59.20}{0.22}
& \ms{1214.3}{14.9}
& \ms{85.18}{0.16}
& \ms{51.14}{0.54}
& \ms{40.66}{0.38} \\

Deterministic top-2
& \textbf{\ms{60.07}{0.27}}
& \textbf{\ms{1293.8}{11.9}}
& \textbf{\ms{86.41}{0.10}}
& \textbf{\ms{55.30}{0.47}}
& \textbf{\ms{42.90}{0.47}} \\

\bottomrule
\end{tabular}
\end{adjustbox}
\end{table}


\end{document}